\documentclass[a4paper]{cas-sc}

\usepackage[numbers]{natbib}

\usepackage{color}
\hypersetup{  % 改变颜色
    colorlinks=true,
    linkcolor=blue,
    filecolor=blue,      
    urlcolor=blue,
    citecolor=cyan,
}
\usepackage{pifont}
\usepackage{graphicx}
\usepackage[justification=centering]{caption}

\usepackage{bbding}
\usepackage{setspace}

\usepackage{bigstrut,multirow,rotating,booktabs}  % 表格插件

\def\tsc#1{\csdef{#1}{\textsc{\lowercase{#1}}\xspace}}
\tsc{WGM}
\tsc{QE}
\tsc{EP}
\tsc{PMS}
\tsc{BEC}
\tsc{DE}
\begin{document}
\begin{sloppypar}

\let\WriteBookmarks\relax
\let\printorcid\relax  % 删除ORCID说明
\def\floatpagepagefraction{1}
\def\textpagefraction{.001}

\shorttitle{BSDP: Brain-inspired Streaming Dual-level Perturbations for Online Open World Object Detection}
\shortauthors{Yu Chen et~al.}

\title[mode = title]{BSDP: Brain-inspired Streaming Dual-level Perturbations for Online Open World Object Detection}

% 作者1
\author[1,2]{Yu Chen}[style=chinese]  % , orcid=0000
% \credit{Conceptualization of this study, Methodology, Software}
\ead{Rain_C1715@shu.edu.cn}  % 邮箱

% 地址
\address[1]{School of Computer Engineering and Science, Shanghai University, Shanghai 200444, China}
\address[2]{Institute of Artificial Intelligence, Shanghai University, Shanghai 200444, China}
\address[3]{Department of Computer and Information Technology, Beijing Jiaotong University, Beijing 100044, China}

% 作者2
\author[1,2]{Liyan Ma}[style=chinese]
\cormark[1]  % 星号角标
\cortext[cor1]{Corresponding author}  % 星号解释
% \ead[url]{https://liyanma-shu.github.io/}  % 主页 带链接
\ead{liyanma@shu.edu.cn}  % 邮箱

% 作者3
\author[3]{Liping Jing}[style=chinese]  % , orcid=0000
% \credit{Conceptualization of this study, Methodology, Software}
\ead{lpjing@bjtu.edu.cn}  % 邮箱

% 作者4
\author[3]{Jian Yu}[style=chinese]  % , orcid=0000
% \credit{Conceptualization of this study, Methodology, Software}
\ead{jianyu@bjtu.edu.cn}  % 邮箱

% 摘要
\begin{abstract}
Humans can easily distinguish the known and unknown categories and can recognize the unknown object by learning it once instead of repeating it many times without forgetting the learned object. Hence, we aim to make deep learning models simulate the way people learn. We refer to such a learning manner as \textbf{O}n\textbf{L}ine \textbf{O}pen \textbf{W}orld \textbf{O}bject \textbf{D}etection(OLOWOD). Existing OWOD approaches pay more attention to the identification of unknown categories, while the incremental learning part is also very important. Besides, some neuroscience research shows that specific noises allow the brain to form new connections and neural pathways which may improve learning speed and efficiency. In this paper, we take the dual-level information of old samples as perturbations on new samples to make the model good at learning new knowledge without forgetting the old knowledge. Therefore, we propose a simple plug-and-play method, called \textbf{B}rain-inspired \textbf{S}treaming \textbf{D}ual-level \textbf{P}erturbations(BSDP), to solve the OLOWOD problem. Specifically, (1) we first calculate the prototypes of previous categories and use the distance between samples and the prototypes as the sample selecting strategy to choose old samples for replay; (2) then take the prototypes as the streaming feature-level perturbations of new samples, so as to improve the plasticity of the model through revisiting the old knowledge; (3) and also use the distribution of the features of the old category samples to generate adversarial data in the form of streams as the data-level perturbations to enhance the robustness of the model to new categories. We empirically evaluate BSDP on PASCAL VOC and MS-COCO, and the excellent results demonstrate the promising performance of our proposed method and learning manner.
\end{abstract}

% 关键词

\maketitle

\doublespacing
% introduction start
\section{Introduction}
\begin{figure}
	\centering
		\includegraphics[scale=.19]{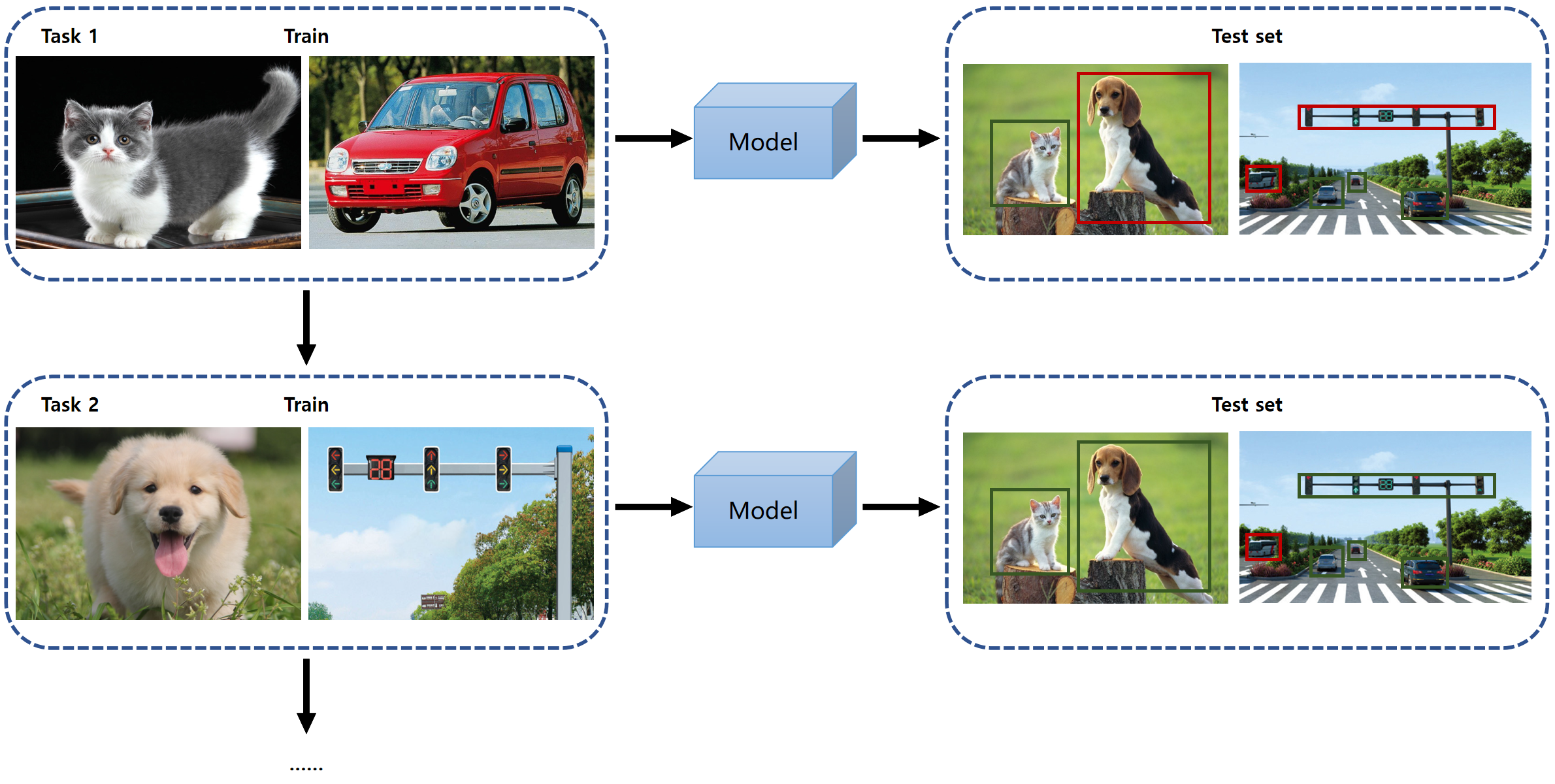}
	\caption{Overview of the OLOWOD problem. The model will see all samples in training sets only once. As the number of tasks increased, the model could recognize more different categories of objects without forgetting previously learned categories. The green boxes on the right indicate objects of known categories, while the red boxes mean that the objects belong to unknown categories.}
	\label{FIG:1}
\end{figure}

The ultimate goal of deep learning research is to enable machines to have the same capabilities of analyzing and learning as humans and to be capable of recognizing data such as texts, images, and sounds. With the continuous development of neural networks, well-trained deep learning models do have better recognition capabilities. However, there is still a big gap between the learning manner of these models and that of humans. First, people clearly know which targets they recognize and which they do not. When confronted with an object, the neurons involved in acquiring target information in the brain are activated and send signals to the primary visual cortex (V1 cortex) in the occipital lobe at the back of the brain to obtain representations. Thus, humans can easily distinguish targets. Even a child who knows only cats and dogs, will not recognize an elephant as a cat or a dog, and reply that he or she does not have the ability to recognize the target. However, traditional deep learning models\cite{MA2023109280, GUO2024110329} recognize all objects in the images of test sets as known objects and do not mark unknown objects. Second, since there are almost 100 billion neurons in the brain, humans have a long-term memory mechanism, which helps to consolidate the memory. Thus, people do not forget the old knowledge they have learned before when they learn new ones. For example, after teaching children to identify elephants, they do not forget what is a cat and what is a dog. However, when fine-tuning the traditional deep learning models with samples from the new categories, the models forget almost all of the previously learned old categories, and the performance of the old categories decreases significantly, which is called the catastrophic forgetting phenomenon\cite{mccloskey1989catastrophic, DBLP:conf/annes/Robins93a}. Finally, people do not need to look at a target repeatedly many times when learning to recognize it. For instance, if a child is given several pictures of an elephant, they only need to learn these pictures once to be able to recognize other pictures of elephants well. However, many existing models require repeated a large number of epochs of training data to achieve the desired recognition accuracy, which is obviously different from the human learning manner. We therefore propose a new evaluation protocol, i.e. \textbf{O}n\textbf{L}ine \textbf{O}pen \textbf{W}orld \textbf{O}bject \textbf{D}etection(OLOWOD), which simulates the human learning manner more realistically. The OLOWOD problem is described in \hyperref[FIG:1]{Figure 1}.

% 图片1

Recently, there have been many works devoted to simulating the real learning manner of people. In open-set task settings, there are categories in the test sets that do not belong to the training sets, and the models need to identify the seen categories normally while identifying the unseen categories as 'unknown'. This experimental protocol strives to allow the model to discriminate between known and unknown categories but is still far from the real-world learning manner. While the open world protocol is closer to the real one than open-set problems. Models not only need to identify unknown categories but also need to have the ability to learn incrementally via replacing the 'unknown' labels with specific categories. Some works have already attempted to work out open world recognition tasks. Bendale et al.\cite{DBLP:conf/cvpr/BendaleB15} proposed NNO to add object categories incrementally while detecting outliers and managing open space risk. However, since open world object detection is quite challenging, it is currently not widely studied. Joseph et al.\cite{DBLP:conf/cvpr/Joseph0KB21} proposed ORE, which was the first model for open world object detection based on Faster R-CNN\cite{DBLP:journals/pami/RenHG017}, to distinguish between known and unknown categories based on auto-labeling unknowns module and energy fitting. Yu et al.\cite{DBLP:conf/icip/YuML0X22} proposed OCPL, which utilized embedding aggregation and prototype learning for the discrimination of known and unknown categories. Compared to ORE\cite{DBLP:conf/cvpr/Joseph0KB21}, OCPL\cite{DBLP:conf/icip/YuML0X22} discards the energy fitting module and avoids the use of energy of unknown categories, and is therefore more consistent with the open world object detection problem.

However, all of these works are only focused on detecting unknown categories and do not pay much attention to investigating incremental learning of unknown categories. Therefore in this work, we focus more on the incremental learning part of the open world object detection problem. Currently, there is a lot of research on incremental learning, some focusing on image classification tasks\cite{DBLP:conf/eccv/LiH16, DBLP:conf/cvpr/RebuffiKSL17, ZHUANG2022108907} and some working on solving object detection tasks\cite{DONG2023109488, DBLP:conf/iccv/ShmelkovSA17, YANG2022108863}. Overall, the incremental learning approach can be divided into two parts based on whether memories are needed or not. Methods that do not require memories rely more on regularization constraints and parameter isolation\cite{DBLP:conf/icml/SerraSMK18} to achieve this. These methods are implemented by tuning the network parameters of the model, which have the advantage that there are no storage and privacy issues, but the performance is biased. While the memory-based approaches, on the other hand, overcome catastrophic forgetting by saving a portion of samples belonging to old categories and fine-tuning the model with these old category exemplars when learning new categories. Compared to memory-free methods, memory-based methods are better for preserving the knowledge of old categories and have better performance. This paper adopts the memory-based approach. We consolidate the knowledge of old categories while learning new categories by using information such as prototypes of old categories and sample distributions in memory, and then use these old category examples to rehearse to better preserve the knowledge of old categories.

Moreover, humans only need to watch a target once when learning to recognize its category, while the training process of traditional models includes dozens or even hundreds of epochs to complete the training. We therefore also change the way the model reads the training data to an online format, thus simulating the human learning manner. Specifically, we use only one epoch of data during training, including exemplars of old categories used for fine-tuning. This is similar to the protocol of some online incremental learning approaches\cite{DBLP:conf/cvpr/Gu0WD22, DBLP:conf/iccv/WangWSG21}. Such a deep model training protocol is more realistic and is the point of studying deep learning models and artificial intelligence. So, how to train the model to better simulate the learning manner of humans? Groen et al.\cite{van2022using} studied the effects of "transcranial random noise stimulation"(tRNS) in a variety of environments, suggesting that some specific noises allow the brain to form new connections and neural pathways that may improve learning speed and efficiency. Therefore, we take advantage of old knowledge as specific noises and propose a novel plug-and-play method called Brain-inspired Streaming Dual-level Perturbations(BSDP), which is suitable for online incremental learning.

The contributions of this paper, in summary, include: 
\begin{itemize} 
\item We first improve the existing evaluation protocol by proposing an online open world experimental baseline, which is more in line with real-world human learning manner. Then, we propose a novel dual-level perturbations method inspired by the human brain’s perception of noise to solve the OLOWOD problem.
\item After selecting samples for replay via a novel prototype-based sample selection strategy, those prototypes of old categories are utilized to disturb features of new category samples as feature-level perturbations. This approach can retain old category features when learning new categories, and better alleviate catastrophic forgetting.
\item We estimate several different types of distributions from the old categories. Then, the data-level perturbations are achieved via disturbing new training data with the noise data sampling from the best-fitted distribution. This extends the new training sets and enhances the robustness of neural networks utilizing adversarial training.
\item We evaluate our approach on standard datasets, such as PASCAL VOC\cite{DBLP:journals/ijcv/EveringhamGWWZ10} and MS-COCO\cite{DBLP:conf/eccv/LinMBHPRDZ14}, and excellent experimental results demonstrate the effectiveness of our work. The proposed new evaluation protocol and baselines pave the way for future studies.
\end{itemize}
% introduction end

% Related work start
\section{Related work}
\subsection{Incremental learning}
Incremental learning(IL) can be classified in many different ways depending on the experimental protocols and implementations. The formulation of incremental learning can be divided into class incremental learning(CIL) and task incremental learning(TIL). According to the implementation of learning, it can be classified into online incremental learning and offline incremental learning. Besides, the methods of incremental learning can be simply grouped into memory-based and memory-free groups based on whether reserve samples of old categories or not.

\subsubsection{CIL vs. TIL}
Task incremental learning(TIL)\cite{DBLP:conf/nips/SunLS0W22, DBLP:conf/cvpr/TiwariKIS22} means that data arriving at different moments are divided into different tasks, and data from the same task can arrive all in one batch. Different tasks do classification or detection independently of each other. Therefore, in the prediction stage, we need to specify the output headers according to the requirements. This also means that doing TIL requires task ID at test time, which is a multi-head setting. Class incremental learning(CIL)\cite{DONG2023109488, SUN2023109561} refers to the fact that data arriving at different moments belong to different classes of the same task. CIL requires the model to perform single-headed outputs and to be able to increase the class number of the outputs. Thus CIL does not require a task ID and is a single-head setting, which means that it is more challenging than TIL.

\subsubsection{Online IL vs. Offline IL}
According to the learning manner, incremental learning can be divided into two categories: online IL\cite{DBLP:conf/cvpr/Gu0WD22, DBLP:conf/iccv/WangWSG21} and offline IL. The traditional deep model learning manner is offline, which means that the data arrives in batches and the model is trained dozens or even hundreds of epochs to get satisfactory results. But this manner is far from the real-world one. In online IL protocol, the data arrives in the form of streams, and models only use each data one epoch instead of repeating multiple epochs. Models are expected to learn new data and new categories in streams and have anti-forgetfulness, which is the manner that the experiments in this paper focus on. Jianren Wang et al.\cite{DBLP:conf/iccv/WangWSG21} proposed an online continual object detection task and provided a novel dataset named OAK. It is similar to OLOWOD to a certain extent, but there are still some differences. They provide short video streams as input training data and there is a very strong correlation between the training images. Besides, they use a small batch to train the model, while OLOWOD is proposed for random image streams with both the batch size and the epoch being 1.

\subsubsection{Memory-based vs. Memory-free}
Memory-free methods are mainly achieved by imposing certain constraints on the parameters of the models. EWC\cite{Kirkpatrick_2017} and MAS\cite{DBLP:conf/eccv/AljundiBERT18} are typical of overcoming catastrophic forgetting by regularizing parameters. There are also methods based on knowledge distillation, which is achieved by transferring knowledge from old tasks to new tasks, such as LwF\cite{DBLP:conf/eccv/LiH16}, iCaRL\cite{DBLP:conf/cvpr/RebuffiKSL17} for incremental classification and ILOD\cite{DBLP:conf/iccv/ShmelkovSA17}, MVCD\cite{YANG2022108863} for incremental object detection, etc.

Memory-based methods preserve a small portion of the samples after learning the current task and replay those reserved samples in the following new tasks. Some methods(MER\cite{DBLP:conf/iclr/RiemerCALRTT19}, Multi-criteria\cite{ZHUANG2022108907}) save real samples of old categories through different sample selection strategies, while others(DGR\cite{DBLP:conf/nips/ShinLKK17}, LGM\cite{DBLP:journals/ijon/RamapuramGK20}) generate pseudo-old samples for replay. This paper adopts the memory-based methods and make better use of prototypes of old category samples by dual-level perturbations to alleviate forgetting.

\subsection{Open-set recognition}\label{2.2}
Some researchers now focus on the open-set task, which is a subset of the open world task. To some degree, our research relates to open-set recognition, which has been studied more. OpenMax\cite{DBLP:conf/cvpr/BendaleB16} replaces Softmax with OpenMax and fits a Weibull model with samples of known categories to determine the probability of classification failure of a closed-set classification model. CROSR\cite{DBLP:conf/cvpr/YoshihashiSKYIN19} enhances the feature representation by jointly classifying and reconstructing the inputs to realize the distinction between known and unknown categories. OpenGAN\cite{DBLP:conf/iccv/KongR21} utilizes generated and real training data to train the discriminator and explores open-set recognition through K-way classification networks. PMAL\cite{DBLP:conf/aaai/LuXLCN22} uses high-quality candidate sample selection and diversity-based filtering to mine useful prototypes for open-set recognition. Zitai Wang et al.\cite{DBLP:conf/nips/WangX00CH22} proposed a novel metric for open-set recognition named OpenAUC and an end-to-end learning method to minimize the OpenAUC risk. In addition to recognition and classification tasks, OSOD\cite{DBLP:conf/wacv/DhamijaGVB20} is the first to formalize the open-set object detection task and to propose the first experimental benchmark for open-set object detection. 
However, methods for open-set recognition or object detection tasks can only distinguish between known categories and unknown categories, which is quite different from human learning manner.

\subsection{Open world object detection}
Compared to the open-set task, the open world task is more relevant to the real world and also is more complex. In the open world protocol, models need to be capable of not only identifying the unknown categories but also identifying the 'unknown' as the specific correct category in subsequent tasks. Open world object detection(OWOD) is more challenging. ORE\cite{DBLP:conf/cvpr/Joseph0KB21} is the first one to formalize the OWOD problem. They obtain an energy model by fitting a Weibull distribution to distinguish between known and unknown categories and store some samples of old categories randomly for replay to overcome forgetting. However, the model uses the ground-truth labels of the unknown categories when fitting the unknown energy. In contrast, OCPL\cite{DBLP:conf/icip/YuML0X22} discards the module of energy fitting and it is based on embedding aggregation and prototype learning for the discrimination of known and unknown categories, which makes OCPL\cite{DBLP:conf/icip/YuML0X22} more in line with the open world protocol. Specifically, OCPL\cite{DBLP:conf/icip/YuML0X22} proposes a prototype branch in the RoI Head and chooses a prototype center with one-hot coding initialization for each category to limit the distribution range of seen categories in the feature space. OCPL\cite{DBLP:conf/icip/YuML0X22} also designs Proposal Embedding Aggregator(PEA) and Embedding Space Compressor(ESC) modules to cluster embedding features of the same category better. However, all of the above methods are offline and focus only on achieving recognition of unknown categories. Compared to OWOD, the OLOWOD problem has more practical significance. If models can be trained in only one epoch to achieve good performance in recognizing known and unknown objects and can be trained incrementally, it can greatly improve industrial efficiency. Besides, models trained in this way are more in line with the human learning manner. Hence, This paper improves OCPL\cite{DBLP:conf/icip/YuML0X22} and achieves good performance in the OLOWOD problem.
% Related work end

% 第三章 start
\section{Proposed methods}
\subsection{Problem formulation}  % 3.1
\subsubsection{Formulation for OWOD}  % 3.1.1 OWOD
We have a series of tasks $T=\{T_1, T_2, ... ,T_N\}$, where $N$ denotes the number of tasks. We consider the set of known categories as $K^t$ for task $T_t$, which is the empty set at the initial stage, and the set of unknown categories as $U^t$. There is no intersection of these two sets, i.e. $K^t \cap U^t = \emptyset $. Each task $T_t$ contains a set of training data $D_{train}^t$ belonging to the categories $C^t=\{c_1,c_2,...,c_n\}$, which is the data accessible to the model at $t$ moments. $n$ is the number of categories included in each task, which we set $n=20$ in the experiments of this paper. The total number of categories $Y$ is obviously the product of the number of tasks $N$ and the number of categories in one single task $n$, i.e. $Y = N \times n$. Before $D_{train}^t$ arrives, the set of categories $C^t$ belongs to the set of unknown categories $U^t$. After the model has completed the task $T_t$, $C^t$ are added to the set of known categories $K^t$ and removed from $U^t$. We train the model for task $T_t$ with the training data $D_{train}^t$ for several epochs, where $t \in \{1,...,N\}$. It means that the model can see the training data several times and can update the weights with the data several times. Data for previous tasks $D_{train}^1,...,D_{train}^{t-1}$ are not available for the task $T_t$. The test data $D_{test}$ is the same for all training tasks $T$, which contains all categories for all tasks.

\subsubsection{Formulation for OLOWOD}  % 3.1.2 OLOWOD
Before describing our BSDP, we formalize the definition of the OLOWOD problem first. In terms of tasks and data splitting, it is the same as OWOD tasks. We have a series of tasks $T=\{T_1,T_2,... ,T_N\}$ and each task $T_t$ contains training data $D_{train}^t$ belonging to the categories $C^t=\{c_1,c_2,... ,c_n\}$. The model can see the training data $D_{train}^1$ for task $T_1$ several times to build a solid base model. But the data $D_{train}^t$ for the incremental task $T_t$ arrives in the form of streams and the model will only use all the data in $D_{train}^t$ for one epoch when training the task $T_t$, where $t \in \{2,...,N\}$. It means that the model can only see the training data once in the incremental steps, which is more challenging but also more realistic significance than OWOD.

To solve the OLOWOD problem, the model is required to be able to identify known categories and to mark unknown categories as 'unknown'. Also, the model should have the ability to learn novel knowledge incrementally in an online manner, which means that it has to learn new categories and not forget old categories while using only one epoch of training data. Specifically, We perform offline training for task $T_1$, simulating that humans need to learn and consolidate knowledge repeatedly when they are teenagers. We train our model for several epochs and the samples can be used more than one time. For subsequent tasks $T_2$, $T_3$, and $T_4$, we perform online training to simulate the learning manner of the young and middle-aged who already have a solid foundation of knowledge. At this stage, samples can be seen only one time and we train the model for only one epoch. 

\subsection{Overall architecture}  % 3.2
% 图片2
\begin{figure*}
	\centering
		\includegraphics[scale=.49]{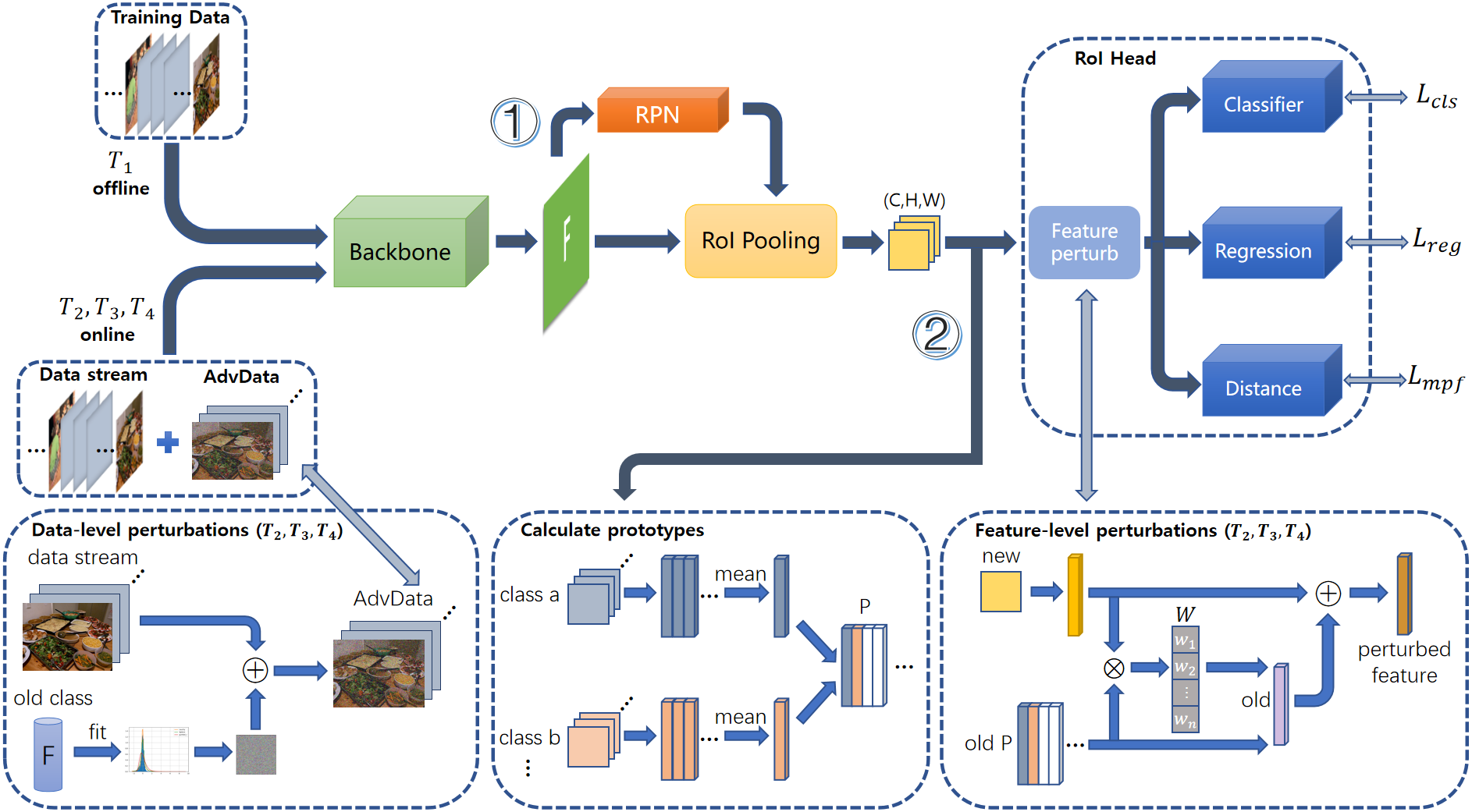}
	\caption{The figure illustrates the overall architecture module of our BSDP. The sign '$\oplus$' represents the addition operation according to certain weights and '$\otimes$' means similarity calculation. 'P' is short for prototype. Prototypes of all seen categories are calculated for sample selection strategies and our dual-level perturbations. In each task, samples are fed into RPN and RoI Pooling to obtain uniform sample features(process \ding{172}), and then into RoI head. After training, the model is used to extract the category features of bounding boxes to calculate prototypes, which is the process \ding{173}. Process \ding{173} is not executed during exemplars replay.}
	\label{FIG:2}
\end{figure*}

The complete architecture of BSDP is shown in \hyperref[FIG:2]{Figure 2}. The training data for each task consists of the original data streams and adversarial data except for the task $T_1$ because there is no old category yet. Our BSDP is a plug-and-play method so it can be combined with ORE\cite{DBLP:conf/cvpr/Joseph0KB21} or OCPL\cite{DBLP:conf/icip/YuML0X22}. Since OCPL\cite{DBLP:conf/icip/YuML0X22} performs better, we implement our approach mainly on it.

In this work, we hope to simulate the learning manner of people in the real world. As is known to all, humans do not have strong learning abilities when they are just born and ready to learn about the world. At this stage, the knowledge reserve of humans is very small and a large amount of data is needed to learn knowledge again and again. Thus, we intend to train task $T_1$ in the offline manner instead of directly in the online manner, because the parameters of the model are completely random and contain no knowledge before training task $T_1$, just like a newborn baby. If training task $T_1$ online directly, the performance will be extremely poor.

\textbf{Training:} As is mentioned above, Task $T_1$ is learned offline. After completion of the task $T_1$, the trained model is used to extract the feature $f_{bbox} \in \mathbb{R}^C$ of the bounding boxes of the samples in $D_{train}^1$, where $C$ is the dimension of the feature vectors and is also the number of channels mentioned in \hyperref[3.3]{Section 3.3}. The obtained sample features $f_{bbox}$ are used to calculate the mean vectors of old categories known as \textbf{prototypes}, which is process \ding{173} in \hyperref[FIG:2]{Figure 2}. In the following training task, we first use the image features of old categories to do data-level perturbations and then we send the obtained adversarial data into the model together with the new coming data to execute process \ding{172}. After the feature vectors $f \in \mathbb{R}^C$ of the same size are obtained by feeding the combined data into the RoI Pooling layer, we use prototypes and $f$ for feature-level perturbations. We then insert the perturbed feature $f_{pert} \in \mathbb{R}^C$ into the classification head, regression head, and distance head of OCPL\cite{DBLP:conf/icip/YuML0X22} and adjust parameters of the neural network through back propagation. After completing the training, we execute process \ding{173} followed task $T_1$. The cycle continues until the last task is completed. After each training task is completed, we replay the selected exemplars to avoid catastrophic forgetting much better.

\textbf{Testing:} In the test phase, the feature perturbation module and process \ding{173} are completely abandoned. After the test data $D_{test}$ is fed into the model, the feature vectors $f$ are obtained by the RPN and RoI Pooling layer. Then we feed $f$ into the classification head, regression head, and distance head to get the final prediction results.

\subsection{Prototype-based sample selection strategy} \label{3.3}  % 3.3
% 图片3
\begin{figure*}
	\centering
		\includegraphics[scale=.56]{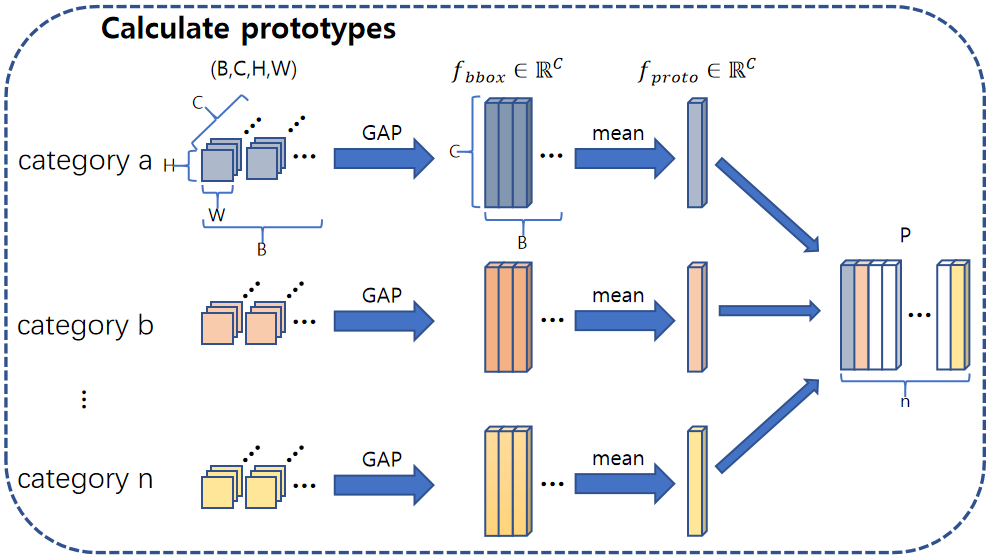}
	\caption{\centering{Description of the calculation of prototypes for our BSDP and dual-level perturbations.}}
	\label{FIG:3}
\end{figure*}

Prototypes of old categories play a crucial role in our BSDP, which can be thought of as memory in human brains. We use the prototypes in our sample selection strategy and apply them to do dual-level perturbations. This subsection details the calculation of prototypes and our prototype-based sample selection strategy.

\textbf{Calculation of prototypes.} For the memory-based incremental learning methods, they select some representative samples(known as exemplars) to replay by specific sample selection strategies. The selected exemplars are a small fraction of the samples of the old categories and are so small that they should be very representative. They are required to include a large amount of knowledge of the old categories so that the model parameters are not too biased toward the new categories during training. Some incremental learning algorithms for image classification have similar conclusions to ours. But compared with image classification, object detection task is more complex. In the object detection task, there are multiple targets in one image and they may belong to different categories. Therefore, we cannot directly extract the feature of the whole image as the feature of a certain category just like what we do in the classification task. After the current task training is completed, we send the training sets to the trained backbone again to get feature maps $F$. Then we feed bounding boxes and $F$ into the RoI Pooling layer to obtain the features of size ($B,C,H,W$) to execute process \ding{173}, where $B$ is the number of targets, $C$ is the number of channels, $H$ and $W$ are the height and width of the features respectively. We then classify these features by their categories, as shown in \hyperref[FIG:3]{Figure 3}. For each category, we do global average pooling(GAP) on the features to obtain $B$ feature vectors $f_{bbox}$ with dimension $C$. Note that the value of $B$ is not necessarily the same for different categories. Finally, we compute the means of all $n$ categories of feature vectors $f_{bbox}$ as prototypes $f_{proto} \in \mathbb{R}^C$ and store them in a matrix $M \in \mathbb{R}^{C \times n}$.

\textbf{Sample selection strategy.} Since the prototypes of old categories contain a lot of implicit previous information, we use the distance between the feature vectors and the prototypes as a measure of whether the sample contains enough old knowledge. The distance $dist$ between the feature $f_{bbox}$ and the prototype $f_{proto}$ of the category $c$ which it belongs to is defined as follows:
% 公式1
\begin{eqnarray}\label{Eq1}
dist= {\parallel}f_{proto}-f_{bbox}{\parallel}_2=\sqrt{\sum^{C}_{k=1}(f_{proto}^k-f_{bbox}^k)^2} \; ,
\end{eqnarray}
where $k$ is the index of the elements in corresponding vectors. The larger the value of distance $dist$ is, the less knowledge the feature contains. Conversely, if the value of distance $dist$ is very small, the feature contains rich knowledge of old categories and is more representative. In our experiments, we follow ORE\cite{DBLP:conf/cvpr/Joseph0KB21} and OCPL\cite{DBLP:conf/icip/YuML0X22} to select 50 exemplars for each category which are the closest to the prototypes to fine-tune the model after training new categories to mitigate catastrophic forgetting.

\subsection{Feature-level perturbations via similarity calculation}\label{3.4}  % 3.4
% 图片4
\begin{figure*}
	\centering
		\includegraphics[scale=.48]{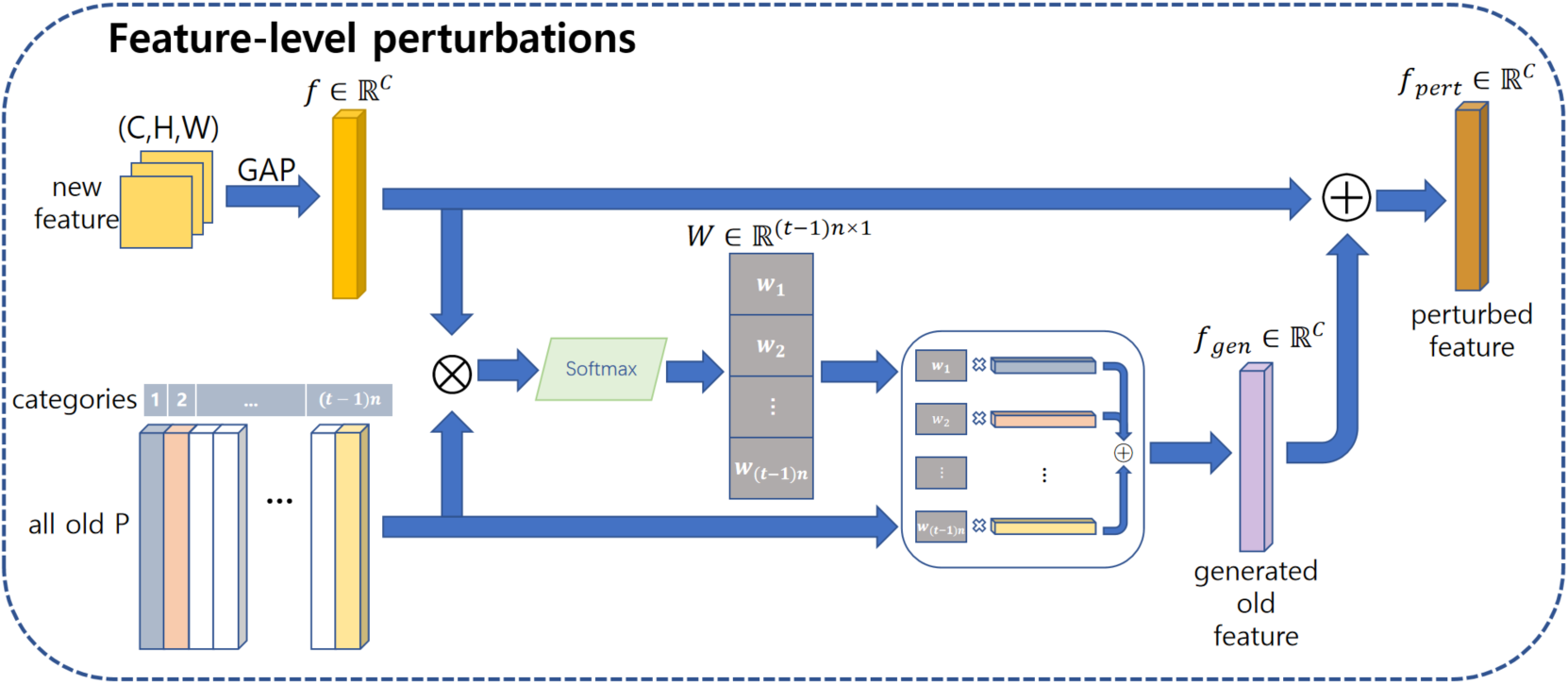}
	\caption{Description of the feature-level perturbations of our BSDP. '$\otimes$' is similarity calculation and '$\oplus$' represents the addition operation according to certain weights. $W$ is the weight vector obtained after similarity calculation.}
	\label{FIG:4}
\end{figure*}

Some memory-based algorithms focus on sample selection strategies. They hope to pick the most representative samples to fine-tune the model to overcome catastrophic forgetting. However, it's far from satisfactory to just fine-tune the model with the chosen exemplars after learning the new task. As mentioned before, some specific noises allow the brain to form new connections and neural pathways to improve learning efficiency. Inspired by human brains, we suggest simulating the learning manner of humans by adding specific perturbations. In fact, the prototypes of old categories can provide more help in overcoming forgetting. Therefore, we propose streaming feature-level perturbations with the help of the prototypes.

Some studies have shown that learning is essentially a process of using old knowledge stored in the brain to interpret new knowledge and then make connections between old and new knowledge. Samara et al.\cite{miller2019functions} pointed out that the brain constantly compares and distinguishes between newly formed sporadic memories and old ones. Therefore, we simulate the way human brains receive new knowledge to make connections between stored old knowledge and new knowledge. Formally, we connect the prototypes of the old categories with the features of new knowledge in a certain proportion as feature-level perturbations. Thus, the perturbed features contain both the correlation and discrimination between the old and new categories. We describe the generation of perturbed features in \hyperref[FIG:4]{Figure 4}. As shown in \hyperref[FIG:2]{Figure 2}, during the training stage of task $T_t$, the feature maps obtained by the RoI Pooling layer of the input images $D_{train}^t$ will be fed into the feature-level perturbation module. We apply global average pooling(GAP) to the feature maps to obtain the feature vectors $f \in \mathbb{R}^C$. At this point, the prototypes of the categories of previous tasks $T_1,..., T_{t-1}$ have been stored in the matrix $M$. As mentioned in the previous works, the model learns a new task resulting in catastrophic forgetting due to lack of training data for the previous tasks. In this paper, we also follow the usual incremental learning settings, where data from previous tasks is not available. Moreover, there is both correlation and discrimination between the features of old and new categories. Therefore, we propose to use the correlation between the features of new samples and the prototypes of old categories to make the model learn discriminative features for each category. Specifically, we respectively calculate the cosine similarity $S_i$ between the feature vectors $f$ of new samples and the prototypes of $(t-1) \times n$ old categories in the matrix $M$:
% 公式2
\begin{eqnarray}\label{Eq2} 
S_i=\frac{f \cdot f_{proto}^i}{{\big|}f{\big|} \cdot {\big|}f_{proto}^i{\big|}}, i \in \{1,2,...,(t-1) \times n\} \;,
\end{eqnarray}

\noindent where $f^i_{proto}$ is the prototype in the matrix $M$ and $i$ is the index of seen categories. Then we send $(t-1) \times n$ similarity values into Softmax to do normalization and get the weight vector $W$:
% 公式3
\begin{eqnarray}\label{Eq3} 
\sigma(S)_i = \frac{e^{S_i}}{\sum^{(t-1) \times n}_{k=1} e^{S_k}} \;,
\end{eqnarray}
% 公式4
\begin{eqnarray}\label{Eq4} 
W=[\sigma_1,\sigma_2,...,\sigma_{(t-1) \times n}] \;.
\end{eqnarray}

The values of each dimension of the weight vector $W \in \mathbb{R}^{(t-1)n \times 1}$ correspond to the feature vectors in the matrix $M \in \mathbb{R}^{C \times (t-1)n}$ one-to-one and represent the correlation between the new sample and all old categories. In order to balance the contributions of all the old categories in the generated features, we multiply the values in the weight vector $W$ with the corresponding prototypes of the old categories in the matrix $M$ and sum up all the results to generate old features $f_{gen}\in \mathbb{R}^C$ which contain both correlation and discrimination between old and new categories. Thus, we have:
% 公式5
\begin{eqnarray}\label{Eq5}
f_{gen}=\sum^{(t-1) \times n}_{i=1} \sigma_i \cdot f^i_{proto} \;,
\end{eqnarray}

\noindent which are perturbations of corresponding features $f$ of new samples. Finally, we add $f_{gen}$ with the features $f$ in a certain proportion to get the perturbed features $f_{pert} \in \mathbb{R}^C$:
% 公式6
\begin{eqnarray}\label{Eq6}
f_{pert} = \gamma \cdot f + (1-\gamma) \cdot f_{gen} \;,
\end{eqnarray}

\noindent where $\gamma$ is the hyper-parameter and is set to be 0.5. The obtained perturbed features $f_{pert}$ contain not only the basic discriminative new category information but also the generalized old knowledge correlated to the features of the new samples.

\subsection{Data-level perturbations via distribution fitting}\label{3.5}  % 3.5
% 图片5
\begin{figure*}
	\centering
		\includegraphics[scale=.19]{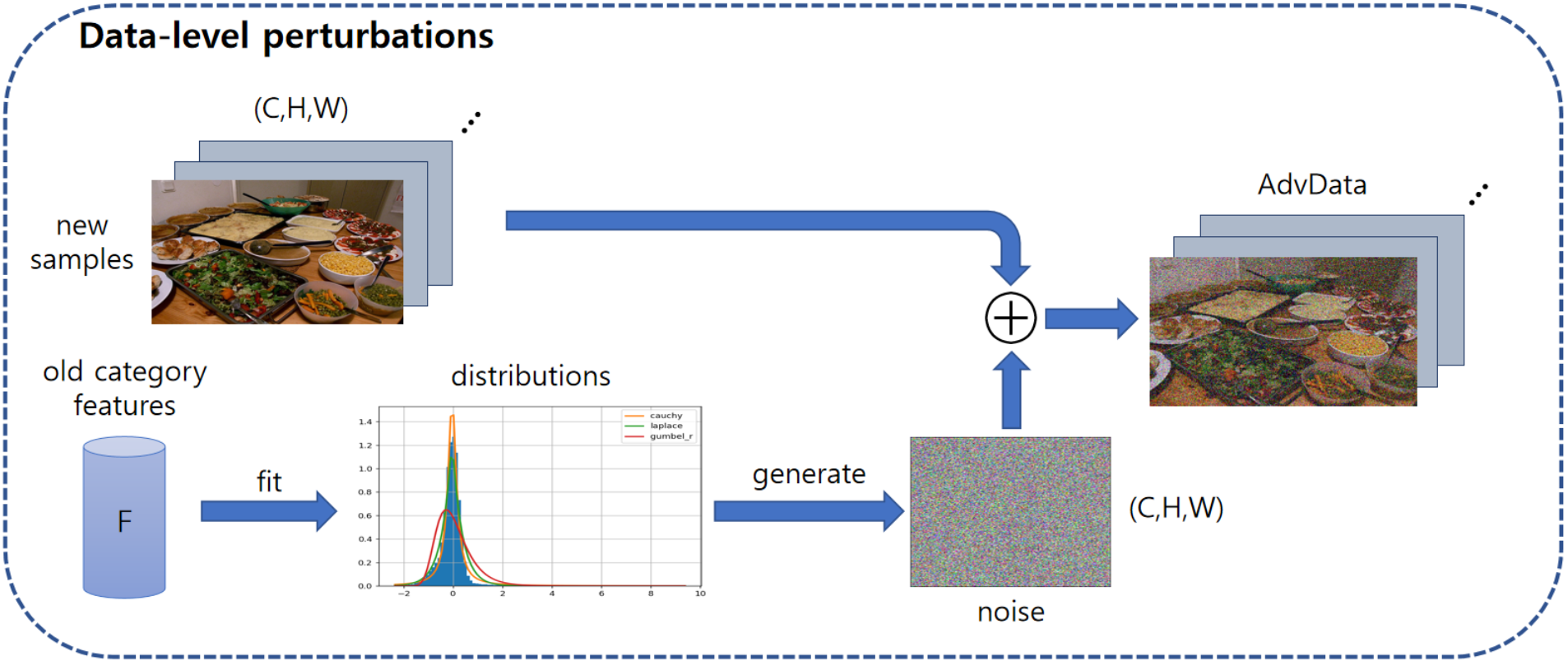}
	\caption{Illustration of the data-level perturbations of our BSDP. $F$ represents the features of samples of each old category stored in addition after model training. We use the features of old categories $F$ to fit the given distributions, which is achieved by using the \textit{Fitter}\cite{Cokelaer2014} library.}
	\label{FIG:5}
\end{figure*}

Although our model can be explored to implicit old knowledge and enhance the discriminative ability with the feature-level perturbation generation module, we wish the model to be able to take full advantage of training samples. It is especially important to make the limited samples provide more knowledge in online learning. McCormick et al.\cite{mccormick1999spontaneous} pointed out that neurons engage in spontaneous neural activity as if neural activity had a mind of its own, which means that even in low-level sensory regions like the visual cortex, neurons encode far more information than they need to fulfill their immediate tasks. Their spontaneous neural activity can be commonly regarded as the noise in brains. Thus, we also add some noises to the model to better simulate what neurons do.

We consider performing data-level perturbations to achieve our goal. Specifically, we generate adversarial data to get more knowledge. Adding some noises to the original samples of new categories, the model can not only acquire more knowledge of new categories but also improve the robustness of the model in online mode. However, the goal of our paper is to expect the model to simulate human learning manner as much as possible. Thus, we propose to generate the adversarial samples by using the statistics of the learned categories rather than using the simple and naive random image transformation or adding common Gaussian noises. Our data-level perturbation generation module is shown in \hyperref[FIG:5]{Figure 5}.

Since we store a fixed number of $k$ sample features for each class of $C_t$ during the training task $T_t(t \ge 1)$, we have $t\times n\times k$ sample features that belong to the $t\times n$ categories before training the task $T_{t+1}$. These features contain a lot of important old category statistics. We use these old category data to fit some types of distributions and calculate the corresponding sum of the squared errors. The smaller the sum of the squared errors is, the better the corresponding distribution fits. We use the Fitter\cite{Cokelaer2014} library to perform the fitting distribution process, which can loop over distributions and find the best parameters to fit the data for each. Then we generate random perturbations from the best fitting distribution as the noise data $N \in \mathbb{R}^{C \times H \times W}$. We use these data-level perturbations to synthesize adversarial data and train the model together with the new coming data. Note that the generated perturbations have the same dimension as the original new category samples, so we can directly add these perturbations to the original images:
% 公式7
\begin{eqnarray}\label{Eq7}
AdvData^{t+1} = D^{t+1}_{train} \oplus N^{t+1} \;,
\end{eqnarray}

\noindent where $N^{t+1} \in \mathbb{R}^{C \times H \times W}$ represent the perturbations generated by the data of the old category samples at step $t+1$. '$\oplus$' means element-wise addition. $AdvData^{t+1}$ are the adversarial samples for task $T_{t+1}$. Then we send the obtained $AdvData ^ {t + 1} $ and new coming samples $D_ {train} ^ {t + 1} $ into the model together to train the task $T_{t + 1}$. It is worth mentioning that we do not perform data-level perturbation generation for every new category sample, but rather randomly select a small portion of the new samples. On the one hand, this can reduce the computing and memory cost. On the other hand, too many perturbed samples will disturb the original data distributions, which leads to the degradation of the feature representations generated by the model as the ablation experiments show. So we only take 1\% of the training samples to perform the data-level perturbation generation in our experiments.

\subsection{Training process}\label{3.6}  % 3.6
% 图片6
\begin{figure*}
	\centering
		\includegraphics[scale=.6]{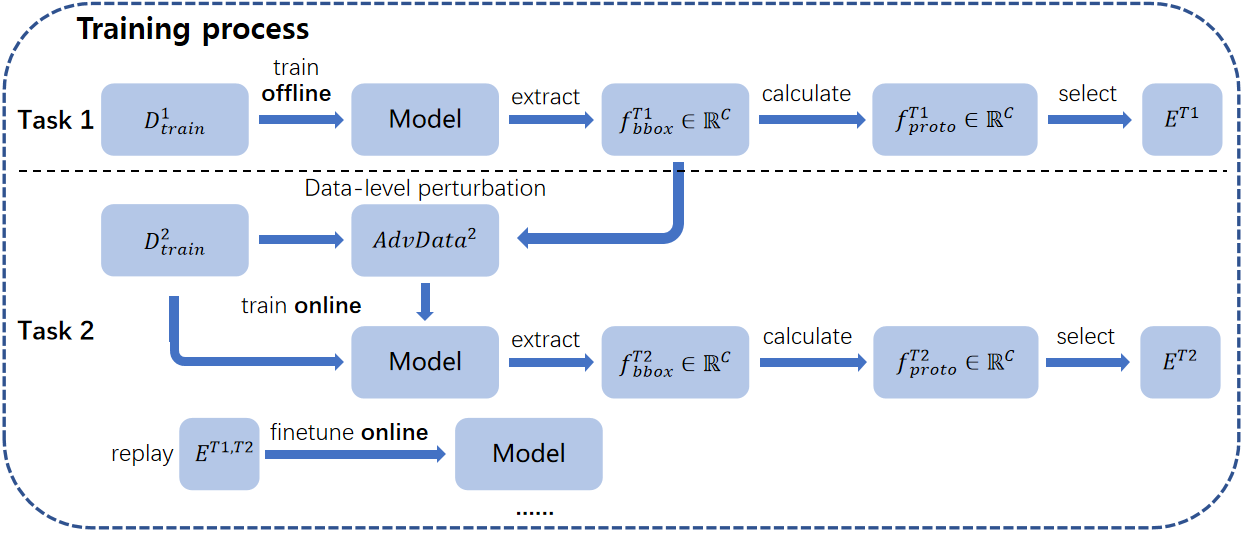}
	\caption{Illustration of the training process of our BSDP. $D$ represents the original training data, while $E$ refers to the selected exemplars for each task. After the training stage of task $T_2$, $T_3$ and $T_4$, we utilize the exemplars of previous and current tasks to fine-tune the model.}
	\label{FIG:6}
\end{figure*}

The overall of the training process is introduced in \hyperref[FIG:6]{Figure 6}. First, we train $T_1$ offline. After training, we utilize the model to extract the feature $f_{bbox} \in \mathbb{R}^{C}$ to calculate the prototypes $f_{proto} \in \mathbb{R}^{C}$ which is utilized to select representative exemplars $E^{T_1}$. Then we train the incremental tasks in an online manner. For tasks $T_2$, $T_3$, and $T_4$, the process of training is quite similar. We first utilize the training data and the features extracted from previous tasks to perform data-level perturbation. Then we send the original data and perturbed data into the model together for online training. We perform the proposed feature-level perturbation in model training.
After that, we obtain the exemplars for the current task. Finally, we use the exemplars of previous and current tasks to fine-tune the model in an online manner. 
% 第三章 end

% 第四章 start
\section{Experiments}
\subsection{Experiment settings}  % 4.1
In this section, we introduce the settings of our experiments, including the datasets, data splits, baselines, metrics, and implementation details.

\subsubsection{Datasets}\label{4.1.1}  % 4.1.1
% 表1
\begin{table}[width=1\textwidth, htbp]
  \centering
  \caption{The description of semantic splits and the number of images for each task for our OLOWOD problem.}
  \setlength{\tabcolsep}{8mm}{
  \scalebox{0.85}{
    \begin{tabular}{c|c|c|c|c}
    \hline
     & \makecell[c]{\textbf{Task 1} \\  \textbf{(base task)}} & \makecell[c]{\textbf{Task 2}} & \makecell[c]{\textbf{Task 3}} & \makecell[c]{\textbf{Task 4}} \\
    \hline
    \makecell[c]{\textbf{Semantic split}} & \makecell[c]{VOC classes} & \makecell[c]{Outdoor,Accessories, \\ Appliance,Truck} & \makecell[c]{Sports, Food} & \makecell[c]{Electronic, Indoor,\\ Kitchen, Furniture} \bigstrut\\
    \hline
    \makecell[c]{\textbf{Training images}} & 16551 & 45520 & 39402 & 40260 \bigstrut\\
    \hline
    \makecell[c]{\textbf{Exemplars}} & 995   & 995   & 994   & 987 \bigstrut\\
    \hline
    \makecell[c]{\textbf{Test images}} & 10246 & 10246 & 10246 & 10246 \bigstrut\\
    \hline
    \end{tabular}}}
  \label{table:1}
\end{table}

In our experiments, we consider the online open world object detection without domain shifts, which contains four tasks as $T=\{T_1, T_2, T_3, T_4\}$. Since our method is based on OCPL\cite{DBLP:conf/icip/YuML0X22}, we follow the data splitting of them. We consider the 20 categories and data in PASCAL VOC\cite{DBLP:journals/ijcv/EveringhamGWWZ10} as task $T_1$. The remaining 60 categories in MS-COCO\cite{DBLP:conf/eccv/LinMBHPRDZ14} are evenly divided into three groups as tasks $T_2$, $T_3$ and $T_4$, as described in OCPL\cite{DBLP:conf/icip/YuML0X22}. Therefore, our incremental learning experimental setups consist of 1 basic task and 3 incremental tasks. The amount of training data and the number of exemplars saved after training contained in each task are shown in \hyperref[table:1]{Table 1}. As mentioned before, we select 50 exemplars for each category in our experiment. While learning the incremental task $T_t(t>1)$, only $D_{train}^{t}$ will be available. All the categories $\{C^i: i \leq t\}$ belong to $\{T_i: i \leq t\}$ will be treated as known and those belong to $\{T_i: i>t\}$ will be considered as unknown. The category sets of different tasks are not intersected, i.e. $C^i \cap C^j = \emptyset, i \neq j$.

\subsubsection{Baselines and metrics}\label{4.1.2}  % 4.1.2
\textbf{Baselines:} We are the first to propose the online open world object detection(OLOWOD) protocol and modify OWOD methods such as ORE\cite{DBLP:conf/cvpr/Joseph0KB21} and OCPL\cite{DBLP:conf/icip/YuML0X22} as our experimental baselines. The original ORE\cite{DBLP:conf/cvpr/Joseph0KB21} and OCPL\cite{DBLP:conf/icip/YuML0X22} are trained offline, which means the models can see all training samples for hundreds of epochs. That is totally different from how people learn. So in this paper, we only train ORE\cite{DBLP:conf/cvpr/Joseph0KB21} and OCPL\cite{DBLP:conf/icip/YuML0X22} for one epoch, which means that the models will see all training samples only once.

\textbf{Metrics:} Followed our baselines, we combine the test split of PASCAL VOC\cite{DBLP:journals/ijcv/EveringhamGWWZ10} and the val split of MS-COCO\cite{DBLP:conf/eccv/LinMBHPRDZ14} as our test set and each task is evaluated using the same test set. For known categories, we still use the mean Average Precision(mAP) at the IoU threshold of 0.5 as the measure of performance. For unknown categories, we follow OCPL\cite{DBLP:conf/icip/YuML0X22} and adopt the Unknown Recall(UR)\cite{DBLP:conf/eccv/BansalSSCD18} rate to measure the model's ability to capture unknown categories. After training on each task except the last one, we use Wilderness Impact(WI)\cite{DBLP:conf/wacv/DhamijaGVB20} to measure the confusion between unknown and known categories:
% 公式8
\begin{eqnarray}\label{Eq8}
Wilderness \: Impact(WI) = \frac{P_K}{P_{K \cup U}} - 1 \;,
\end{eqnarray}

\noindent where $P_K$ represents the accuracy of the model evaluated on known categories and $P_{K \cup U}$ refers to the accuracy when evaluated on known and unknown categories. In addition, we use the Absolute Open-Set Error(A-OSE)\cite{DBLP:conf/icra/MillerNDS18} to report the number of unknown category objects that are incorrectly detected as any known category after learning a task.

\subsubsection{Implementation details}\label{4.1.3}  % 4.1.3
All the models in our experiments adopt Faster R-CNN\cite{DBLP:journals/pami/RenHG017} and use ResNet-50\cite{DBLP:conf/cvpr/HeZRS16} as the backbone. The code is implemented in PyTorch and we use the Detectron2\cite{wu2019detectron2} framework. For task $T_1$, we use the same offline training manner as OCPL\cite{DBLP:conf/icip/YuML0X22} to get a base model and do not perform dual-level perturbations because there are no old categories yet. In a series of subsequent tasks $T_i(i>1)$, training data is fed into the model in the form of streams. We use the prototypes and features saved from all previous tasks to help the model train better in an online manner. For the data-level perturbation module, we store 80 features as a group and saved 20 groups of features for each old category to fit the distributions in our experiments. We generate data-level perturbations for 1\% training samples. For our prototype-based sample selection strategy, we empirically choose to store 50 samples for each category that are the closest to the old prototypes.

\subsection{Results and discussion}  % 4.2
In this section, we compare our BSDP with the proposed baselines and perform ablation experiments on our approach to better analyze its effectiveness.

\subsubsection{Comparative results}\label{4.2.1}  % 4.2.1
% 表2
\begin{table*}[width=1\textwidth, htbp]
  \centering
  \caption{Here we show the experimental results of our proposed BSDP and modified baselines in OLOWOD protocol. The 'previous', 'current', and 'both' represent mAPs of old categories, the categories of the current task, and all known categories respectively. (↓) indicates that the lower the value is, the better the model performs, while (↑) means higher is better. The results in grey are based on ORE\cite{DBLP:conf/cvpr/Joseph0KB21}, while the results in white are OCPL\cite{DBLP:conf/icip/YuML0X22} based. The best results have been bolded in the table.}
  \setlength{\tabcolsep}{2mm}{
  \resizebox{\textwidth}{1.8cm}{
  \scalebox{0.63}{
    \begin{tabular}{l|c|c|c|c|c|c|ccc|c|c|c|ccc|c|ccc}
    \hline
    Task IDs & \multicolumn{4}{c|}{Task 1}   & \multicolumn{6}{c|}{Task 2}                   & \multicolumn{6}{c|}{Task 3}                   & \multicolumn{3}{c}{Task 4} \bigstrut\\
    \hline
          & WI    & A-OSE & mAP(↑) & UR    & WI    & A-OSE & \multicolumn{3}{c|}{mAP(↑)} & UR    & WI    & A-OSE & \multicolumn{3}{c|}{mAP(↑)} & UR    & \multicolumn{3}{c}{mAP(↑)} \bigstrut\\
\cline{4-4}\cline{8-10}\cline{14-16}\cline{18-20}          & (↓)   & (↓)   & current & (↑)   & (↓)   & (↓)   & previous & current & both  & (↑)   & (↓)   & (↓)   & previous & current & both  & (↑)   & previous & current & both \bigstrut\\
    \hline
    \rowcolor[rgb]{ .949,  .949,  .949} ORE*  & 0.0540 & 12057 & 56.11 & 5.68  & 0.0293 & 11270 & 51.77 & 26.44 & 39.10 & 3.64  & 0.0203 & 9121  & 37.85 & 12.98 & 29.56 & 3.90  & \multicolumn{1}{c|}{29.49} & 13.30 & 25.44 \bigstrut\\
    \hline
    OCPL* & 0.0427 & 5520  & 56.15 & 8.25  & 0.0229 & 5875  & 51.46 & 26.70 & 39.08 & 7.29  & 0.0156 & 4744  & 38.72 & 14.82 & 30.76 & 11.77 & \multicolumn{1}{c|}{30.67} & 14.23 & 26.56 \bigstrut\\
    \hline
    \rowcolor[rgb]{ .949,  .949,  .949} ORE*-online & 0.0540 & 12057 & 56.11 & 5.68  & 0.0367 & 12202 & 50.69 & 21.16 & 35.92 & 3.98  & 0.0186 & 8808  & 36.36 & 11.84 & 28.19 & 2.62  & \multicolumn{1}{c|}{28.05} & 11.74 & 23.97 \bigstrut\\
    \hline
    OCPL*-online & 0.0427 & 5520  & 56.15 & 8.25  & \textbf{0.0224} & 5817  & 51.22 & 23.40 & 37.31 & \textbf{7.76} & \textbf{0.0162} & \textbf{3802} & 37.89 & 12.51 & 29.43 & 9.40  & \multicolumn{1}{c|}{28.98} & 12.27 & 24.81 \bigstrut\\
    \hline
    \rowcolor[rgb]{ .949,  .949,  .949} BSDP(ORE*) & 0.0540 & 12057 & 56.11 & 5.68  & 0.0365 & 13622 & 52.36 & 21.96 & 37.16 & 3.52  & 0.0238 & 7752  & 38.73 & 11.31 & 29.60 & 3.37  & 29.83 & 11.21 & 25.18 \bigstrut\\
    \hline
    BSDP(OCPL*) & \textbf{0.0427} & \textbf{5520} & \textbf{56.15} & \textbf{8.25} & 0.0243 & \textbf{5386} & \textbf{53.28} & \textbf{23.73} & \textbf{38.50} & 7.44  & 0.0168 & 4308  & \textbf{39.31} & \textbf{13.37} & \textbf{30.66} & \textbf{10.30} & \textbf{30.73} & \textbf{12.88} & \textbf{26.26} \bigstrut\\
    \hline
    \end{tabular}}}}
  \label{table:2}
\end{table*}

\hyperref[table:2]{Table 2} shows the performance of our proposed BSDP compared with other baselines. 'ORE*' and OCPL*' indicate the offline learning method ORE\cite{DBLP:conf/cvpr/Joseph0KB21} and OCPL\cite{DBLP:conf/icip/YuML0X22}, respectively. As mentioned in OCPL\cite{DBLP:conf/icip/YuML0X22}, ORE\cite{DBLP:conf/cvpr/Joseph0KB21} is the first work for open world object detection. Since its energy-based unknown identification module does not fit the OWOD scenario, we discard this module and represent it as 'ORE*' for fair comparison. We also modify ORE* to follow the OLOWOD setting as one of our baselines, represented by 'ORE*-online'. 'OCPL*' represents an OCPL\cite{DBLP:conf/icip/YuML0X22} implementation without the cosine similarity-based classifier, which helps little or even reduces model performance in the tasks trained in the online mode. 'OCPL*-online' indicates the OCPL\cite{DBLP:conf/icip/YuML0X22} follows the OLOWOD setting. This is our second baseline. Therefore, the results of the first two rows in \hyperref[table:2]{Table 2} are obtained by offline training and can be treated as the upper bounds, while the results of row 3 and row 4 are obtained by online training. Since the task $T_1$ trains 20 categories offline, the model is well initialized, thus contributing to the subsequent performance improvement. When investigating the effect of the category number of the base task, we find that a smaller number of base categories in task $T_1$ results in a decrease in overall performance. Similar to the algorithms utilizing pretrained models, a better base model will be beneficial to the following tasks. We have performed some experiments and found this phenomenon. In future work, we will further explore the impact of the base task.

As shown in row 5 and row 6 of \hyperref[table:2]{Table 2}, our BSDP approach is highly competitive and outperforms ORE\*-online and OCPL\*-online obviously in almost all the metrics. When focusing on the mAPs on subsequent incremental tasks, it can be seen that the mAPs of both the old and new categories have improved by nearly 1\%, and some have improved by as much as 2\%. Since ORE* and OCPL* are trained offline, the results in row 1 and row 2 can be treated as the upper bound. When we compare the results on row 1/3/5, we can see that most of the mAPs of ORE*-based BSDP have been improved. The mAPs of the known categories of tasks $T_2, T_3$ and $T_4$ increased by 1.24\%, 1.41\%, and 1.21\% respectively. Since ORE* has little ability to handle unknown categories, all metrics of the unknown detection are much lower than OCPL* in the case of online incremental learning. When we compare the results on row 2/4/6, we can see that our approach also outperforms OCPL*. The mAPs of tasks $T_2,T_3$ and $T_4$ improved by 1.19\%, 1.23\% and 1.45\% respectively. Since OCPL* can efficiently handle unknown categories, it helps to improve the performance of known categories by avoiding the misclassification of unknown categories. 

It is worth mentioning that our BSDP methods, whether based on ORE* or OCPL*, perform even better on previous categories than those trained offline(e.g. 53.28\% vs. 51.46\%, 39.31\% vs. 38.72\% and 30.73\% vs. 30.67\% for OCPL*). The accuracies on 'both' are also close to those trained in an offline manner. It proves that our OLOWOD protocol is more realistic and our BSDP has promising future explorations and applications.

When focusing on the metrics for unknown detection, it can be found that OCPL*-based methods outperform ORE*-based ones in all metrics. Compared the results of row 2 with row 4, OCPL* performs well in both offline and online tasks on unknown detection, which also proves the generalization of OCPL*. Compared the results of row 4 with row 6, our OCPL*-based BSDP effectively improves all mAPs on all incremental tasks, not affecting the performance of unknown detection and even improving A-OSE for task $T_2$ and UR for task $T_3$.

Compared the performance on different tasks, it is obvious that there is a trend of change. As the task continues to be trained, the performance of the model learning the category of the current task always decreases, as well as the performance of the previous task. However, the performance of unknown detection increases. On the one hand, the weights of models already make sense when training new tasks. Learning new tasks is bound to adjust the weights of the models, which inevitably leads to catastrophic forgetting. On the other hand, the significance of the model weights means that the weight values cannot be changed freely when learning new tasks, which may cause mediocre performance for the current task. This is the Stability-Plasticity Dilemma for Artificial Intelligence. As for the performance of unknown detection, it is because as the task continues to learn, the number of unknown categories becomes less and less, just as people grow up with fewer and fewer objects they do not recognize.

\subsubsection{Comparison for CIL}  % 4.2.2
We also perform experiments on class incremental learning and we compare our BSDP with other incremental detection methods. We follow ORE\cite{DBLP:conf/cvpr/Joseph0KB21} to use the standard protocol used in the iOD domain to evaluate our BSDP, where a group of categories (10,5 and the last one) from PASCAL VOC\cite{DBLP:journals/ijcv/EveringhamGWWZ10} are incrementally learned by a detector trained on the remaining set of categories. The experiments on class incremental learning are performed offline. We show the mAPs of each category after training on the incremental task in \hyperref[table:3]{Table 3}. The results of the methods except OCPL* and BSDP are copied from their original paper.
% 表3
\begin{table*}[width=1\textwidth, htbp]
  \centering
  \caption{We compare our BSDP with other incremental object detectors in three different settings. 10, 5, and the last class from PASCAL VOC\cite{DBLP:journals/ijcv/EveringhamGWWZ10} are introduced to a detector trained on 10, 15, and 19 classes respectively (shown in \colorbox[rgb]{ 1,  .949,  .8}{yellow} background). The experiments are performed offline.}
  \scalebox{0.64}{
    \begin{tabular}{l|c|c|c|c|c|c|c|c|c|c|c|c|c|c|c|c|c|c|c|c|c}
    \hline
    \rowcolor[rgb]{ .906,  .902,  .902} \textcolor[rgb]{ .329,  .51,  .208}{\textbf{10+10 settings}} & \textbf{aero} & \textbf{cycle} & \textbf{bird} & \textbf{boat} & \textbf{bottle} & \textbf{bus} & \textbf{car} & \textbf{cat} & \textbf{chair} & \textbf{cow} & \textbf{table} & \textbf{dog} & \textbf{horse} & \textbf{bike} & \textbf{person} & \textbf{plant} & \textbf{sheep} & \textbf{sofa} & \textbf{train} & \textbf{tv} & \textbf{mAP} \bigstrut\\
    \hline
    ILOD\cite{DBLP:conf/iccv/ShmelkovSA17} & 69.9  & 70.4  & 69.4  & 54.3  & 48.0  & 68.7  & 78.9  & 68.4  & 45.5  & 58.1  & \cellcolor[rgb]{ 1,  .949,  .8}59.7 & \cellcolor[rgb]{ 1,  .949,  .8}72.7 & \cellcolor[rgb]{ 1,  .949,  .8}73.5 & \cellcolor[rgb]{ 1,  .949,  .8}73.2 & \cellcolor[rgb]{ 1,  .949,  .8}66.3 & \cellcolor[rgb]{ 1,  .949,  .8}29.5 & \cellcolor[rgb]{ 1,  .949,  .8}63.4 & \cellcolor[rgb]{ 1,  .949,  .8}61.6 & \cellcolor[rgb]{ 1,  .949,  .8}69.3 & \cellcolor[rgb]{ 1,  .949,  .8}62.2 & 63.1 \bigstrut\\
    \hline
    Faster ILOD\cite{DBLP:journals/prl/PengZL20} & 72.8  & 75.7  & 71.2  & 60.5  & 61.7  & 70.4  & 83.3  & 76.6  & 53.1  & 72.3  & \cellcolor[rgb]{ 1,  .949,  .8}36.7 & \cellcolor[rgb]{ 1,  .949,  .8}70.9 & \cellcolor[rgb]{ 1,  .949,  .8}66.8 & \cellcolor[rgb]{ 1,  .949,  .8}67.6 & \cellcolor[rgb]{ 1,  .949,  .8}66.1 & \cellcolor[rgb]{ 1,  .949,  .8}24.7 & \cellcolor[rgb]{ 1,  .949,  .8}63.1 & \cellcolor[rgb]{ 1,  .949,  .8}48.1 & \cellcolor[rgb]{ 1,  .949,  .8}57.1 & \cellcolor[rgb]{ 1,  .949,  .8}43.6 & 62.2 \bigstrut\\
    \hline
    iOD\cite{DBLP:journals/pami/JosephRKKB22}   & 76.0  & 74.6  & 67.5  & 55.9  & 57.6  & 75.1  & 85.4  & 77.0  & 43.7  & 70.8  & \cellcolor[rgb]{ 1,  .949,  .8}60.1 & \cellcolor[rgb]{ 1,  .949,  .8}66.4 & \cellcolor[rgb]{ 1,  .949,  .8}76.0 & \cellcolor[rgb]{ 1,  .949,  .8}72.6 & \cellcolor[rgb]{ 1,  .949,  .8}74.6 & \cellcolor[rgb]{ 1,  .949,  .8}39.7 & \cellcolor[rgb]{ 1,  .949,  .8}64.0 & \cellcolor[rgb]{ 1,  .949,  .8}60.2 & \cellcolor[rgb]{ 1,  .949,  .8}68.5 & \cellcolor[rgb]{ 1,  .949,  .8}60.5 & 66.3 \bigstrut\\
    \hline
    ORE\cite{DBLP:conf/cvpr/Joseph0KB21}  & 63.5  & 70.9  & 58.9  & 42.9  & 34.1  & 76.2  & 80.7  & 76.3  & 34.1  & 66.1  & \cellcolor[rgb]{ 1,  .949,  .8}56.1 & \cellcolor[rgb]{ 1,  .949,  .8}70.4 & \cellcolor[rgb]{ 1,  .949,  .8}80.2 & \cellcolor[rgb]{ 1,  .949,  .8}72.3 & \cellcolor[rgb]{ 1,  .949,  .8}81.8 & \cellcolor[rgb]{ 1,  .949,  .8}42.7 & \cellcolor[rgb]{ 1,  .949,  .8}71.6 & \cellcolor[rgb]{ 1,  .949,  .8}68.1 & \cellcolor[rgb]{ 1,  .949,  .8}77.0 & \cellcolor[rgb]{ 1,  .949,  .8}67.7 & 64.6 \bigstrut\\
    \hline
    OCPL* & 84.8  & 83.2  & 73.8  & 62.9  & 65.5  & 82.5  & 88.8  & 84.4  & 54.2  & 60.0  & \cellcolor[rgb]{ 1,  .949,  .8}46.1 & \cellcolor[rgb]{ 1,  .949,  .8}50.8 & \cellcolor[rgb]{ 1,  .949,  .8}65.2 & \cellcolor[rgb]{ 1,  .949,  .8}66.0 & \cellcolor[rgb]{ 1,  .949,  .8}71.9 & \cellcolor[rgb]{ 1,  .949,  .8}21.6 & \cellcolor[rgb]{ 1,  .949,  .8}49.8 & \cellcolor[rgb]{ 1,  .949,  .8}54.6 & \cellcolor[rgb]{ 1,  .949,  .8}68.5 & \cellcolor[rgb]{ 1,  .949,  .8}46.3 & 64.3 \bigstrut\\
    \hline
    BSDP(ORE*) & 86.1  & 85.1  & 74.5  & 63.8  & 66.5  & 82.8  & 90.7  & 78.9  & 53.1  & 60.9  & \cellcolor[rgb]{ 1,  .949,  .8}49.3 & \cellcolor[rgb]{ 1,  .949,  .8}66.4 & \cellcolor[rgb]{ 1,  .949,  .8}68.6 & \cellcolor[rgb]{ 1,  .949,  .8}72.1 & \cellcolor[rgb]{ 1,  .949,  .8}76.6 & \cellcolor[rgb]{ 1,  .949,  .8}32.0 & \cellcolor[rgb]{ 1,  .949,  .8}60.5 & \cellcolor[rgb]{ 1,  .949,  .8}58.1 & \cellcolor[rgb]{ 1,  .949,  .8}65.8 & \cellcolor[rgb]{ 1,  .949,  .8}58.7 & 67.6 \bigstrut\\
    \hline
    BSDP(OCPL*) & 86.8  & 84.7  & 74.3  & 62.8  & 64.4  & 83.9  & 90.3  & 82.5  & 54.3  & 75.5  & \cellcolor[rgb]{ 1,  .949,  .8}48.2 & \cellcolor[rgb]{ 1,  .949,  .8}66.6 & \cellcolor[rgb]{ 1,  .949,  .8}72.1 & \cellcolor[rgb]{ 1,  .949,  .8}72.8 & \cellcolor[rgb]{ 1,  .949,  .8}77.3 & \cellcolor[rgb]{ 1,  .949,  .8}32.9 & \cellcolor[rgb]{ 1,  .949,  .8}65.4 & \cellcolor[rgb]{ 1,  .949,  .8}60.8 & \cellcolor[rgb]{ 1,  .949,  .8}69.7 & \cellcolor[rgb]{ 1,  .949,  .8}58.9 & \textbf{69.2} \bigstrut\\
    \hline
    \multicolumn{1}{r}{} & \multicolumn{1}{r}{} & \multicolumn{1}{r}{} & \multicolumn{1}{r}{} & \multicolumn{1}{r}{} & \multicolumn{1}{r}{} & \multicolumn{1}{r}{} & \multicolumn{1}{r}{} & \multicolumn{1}{r}{} & \multicolumn{1}{r}{} & \multicolumn{1}{r}{} & \multicolumn{1}{r}{} & \multicolumn{1}{r}{} & \multicolumn{1}{r}{} & \multicolumn{1}{r}{} & \multicolumn{1}{r}{} & \multicolumn{1}{r}{} & \multicolumn{1}{r}{} & \multicolumn{1}{r}{} & \multicolumn{1}{r}{} & \multicolumn{1}{r}{} &  \bigstrut\\
    \hline
    \rowcolor[rgb]{ .906,  .902,  .902} \textcolor[rgb]{ .329,  .51,  .208}{\textbf{15+5 settings}} & \textbf{aero} & \textbf{cycle} & \textbf{bird} & \textbf{boat} & \textbf{bottle} & \textbf{bus} & \textbf{car} & \textbf{cat} & \textbf{chair} & \textbf{cow} & \textbf{table} & \textbf{dog} & \textbf{horse} & \textbf{bike} & \textbf{person} & \textbf{plant} & \textbf{sheep} & \textbf{sofa} & \textbf{train} & \textbf{tv} & \textbf{mAP} \bigstrut\\
    \hline
    ILOD\cite{DBLP:conf/iccv/ShmelkovSA17} & 70.5  & 79.2  & 68.8  & 59.1  & 53.2  & 75.4  & 79.4  & 78.8  & 46.6  & 59.4  & 59.0  & 75.8  & 71.8  & 78.6  & 69.6  & \cellcolor[rgb]{ 1,  .949,  .8}33.7 & \cellcolor[rgb]{ 1,  .949,  .8}61.5 & \cellcolor[rgb]{ 1,  .949,  .8}63.1 & \cellcolor[rgb]{ 1,  .949,  .8}71.7 & \cellcolor[rgb]{ 1,  .949,  .8}62.2 & 65.9 \bigstrut\\
    \hline
    Faster ILOD\cite{DBLP:journals/prl/PengZL20} & 66.5  & 78.1  & 71.8  & 54.6  & 61.4  & 68.4  & 82.6  & 82.7  & 52.1  & 74.3  & 63.1  & 78.6  & 80.5  & 78.4  & 80.4  & \cellcolor[rgb]{ 1,  .949,  .8}36.7 & \cellcolor[rgb]{ 1,  .949,  .8}61.7 & \cellcolor[rgb]{ 1,  .949,  .8}59.3 & \cellcolor[rgb]{ 1,  .949,  .8}67.9 & \cellcolor[rgb]{ 1,  .949,  .8}59.1 & 67.9 \bigstrut\\
    \hline
    iOD\cite{DBLP:journals/pami/JosephRKKB22}   & 78.4  & 79.7  & 66.9  & 54.8  & 56.2  & 77.7  & 84.6  & 79.1  & 47.7  & 75.0  & 61.8  & 74.7  & 81.6  & 77.5  & 80.2  & \cellcolor[rgb]{ 1,  .949,  .8}37.8 & \cellcolor[rgb]{ 1,  .949,  .8}58.0 & \cellcolor[rgb]{ 1,  .949,  .8}54.6 & \cellcolor[rgb]{ 1,  .949,  .8}73.0 & \cellcolor[rgb]{ 1,  .949,  .8}56.1 & 67.8 \bigstrut\\
    \hline
    ORE\cite{DBLP:conf/cvpr/Joseph0KB21}  & 75.4  & 81.0  & 67.1  & 51.9  & 55.7  & 77.2  & 85.6  & 81.7  & 46.1  & 76.2  & 55.4  & 76.7  & 86.2  & 78.5  & 82.1  & \cellcolor[rgb]{ 1,  .949,  .8}32.8 & \cellcolor[rgb]{ 1,  .949,  .8}63.6 & \cellcolor[rgb]{ 1,  .949,  .8}54.7 & \cellcolor[rgb]{ 1,  .949,  .8}77.7 & \cellcolor[rgb]{ 1,  .949,  .8}64.6 & 68.5 \bigstrut\\
    \hline
    OCPL* & 81.5  & 88.5  & 77.3  & 65.2  & 73.2  & 81.3  & 90.9  & 89.0  & 61.2  & 80.3  & 75.5  & 81.2  & 89.2  & 84.4  & 83.3  & \cellcolor[rgb]{ 1,  .949,  .8}19.1 & \cellcolor[rgb]{ 1,  .949,  .8}25.2 & \cellcolor[rgb]{ 1,  .949,  .8}24.0 & \cellcolor[rgb]{ 1,  .949,  .8}65.1 & \cellcolor[rgb]{ 1,  .949,  .8}36.8 & 69.5 \bigstrut\\
    \hline
    BSDP(ORE*) & 86.6  & 88.9  & 79.0  & 67.4  & 71.5  & 81.3  & 91.1  & 90.0  & 63.7  & 79.4  & 73.8  & 83.4  & 89.3  & 87.7  & 87.3  & \cellcolor[rgb]{ 1,  .949,  .8}31.3 & \cellcolor[rgb]{ 1,  .949,  .8}64.1 & \cellcolor[rgb]{ 1,  .949,  .8}57.4 & \cellcolor[rgb]{ 1,  .949,  .8}74.0 & \cellcolor[rgb]{ 1,  .949,  .8}53.8 & 74.1 \bigstrut\\
    \hline
    BSDP(OCPL*) & 84.2  & 89.5  & 78.5  & 66.7  & 72.3  & 84.7  & 91.5  & 88.5  & 61.9  & 81.4  & 72.9  & 84.4  & 89.7  & 87.9  & 87.6  & \cellcolor[rgb]{ 1,  .949,  .8}30.3 & \cellcolor[rgb]{ 1,  .949,  .8}64.8 & \cellcolor[rgb]{ 1,  .949,  .8}61.1 & \cellcolor[rgb]{ 1,  .949,  .8}75.7 & \cellcolor[rgb]{ 1,  .949,  .8}54.0 & \textbf{75.5} \bigstrut\\
    \hline
    \multicolumn{1}{r}{} & \multicolumn{1}{r}{} & \multicolumn{1}{r}{} & \multicolumn{1}{r}{} & \multicolumn{1}{r}{} & \multicolumn{1}{r}{} & \multicolumn{1}{r}{} & \multicolumn{1}{r}{} & \multicolumn{1}{r}{} & \multicolumn{1}{r}{} & \multicolumn{1}{r}{} & \multicolumn{1}{r}{} & \multicolumn{1}{r}{} & \multicolumn{1}{r}{} & \multicolumn{1}{r}{} & \multicolumn{1}{r}{} & \multicolumn{1}{r}{} & \multicolumn{1}{r}{} & \multicolumn{1}{r}{} & \multicolumn{1}{r}{} & \multicolumn{1}{r}{} &  \bigstrut\\
    \hline
    \rowcolor[rgb]{ .906,  .902,  .902} \textcolor[rgb]{ .329,  .51,  .208}{\textbf{19+1 settings}} & \textbf{aero} & \textbf{cycle} & \textbf{bird} & \textbf{boat} & \textbf{bottle} & \textbf{bus} & \textbf{car} & \textbf{cat} & \textbf{chair} & \textbf{cow} & \textbf{table} & \textbf{dog} & \textbf{horse} & \textbf{bike} & \textbf{person} & \textbf{plant} & \textbf{sheep} & \textbf{sofa} & \textbf{train} & \textbf{tv} & \textbf{mAP} \bigstrut\\
    \hline
    ILOD\cite{DBLP:conf/iccv/ShmelkovSA17} & 69.4  & 79.3  & 69.5  & 57.4  & 45.4  & 78.4  & 79.1  & 80.5  & 45.7  & 76.3  & 64.8  & 77.2  & 80.8  & 77.5  & 70.1  & 42.3  & 67.5  & 64.4  & 76.7  & \cellcolor[rgb]{ 1,  .949,  .8}62.7 & 68.3 \bigstrut\\
    \hline
    Faster ILOD\cite{DBLP:journals/prl/PengZL20} & 64.2  & 74.7  & 73.2  & 55.5  & 53.7  & 70.8  & 82.9  & 82.6  & 51.6  & 79.7  & 58.7  & 78.8  & 81.8  & 75.3  & 77.4  & 43.1  & 73.8  & 61.7  & 69.8  & \cellcolor[rgb]{ 1,  .949,  .8}61.1 & 68.6 \bigstrut\\
    \hline
    iOD\cite{DBLP:journals/pami/JosephRKKB22}   & 78.2  & 77.5  & 69.4  & 55.0  & 56.0  & 78.4  & 84.2  & 79.2  & 46.6  & 79.0  & 63.2  & 78.5  & 82.7  & 79.1  & 79.9  & 44.1  & 73.2  & 66.3  & 76.4  & \cellcolor[rgb]{ 1,  .949,  .8}57.6 & 70.2 \bigstrut\\
    \hline
    ORE\cite{DBLP:conf/cvpr/Joseph0KB21}  & 67.3  & 76.8  & 60.0  & 48.4  & 58.8  & 81.1  & 86.5  & 75.8  & 41.5  & 79.6  & 54.6  & 72.8  & 85.9  & 81.7  & 82.4  & 44.8  & 75.8  & 68.2  & 75.7  & \cellcolor[rgb]{ 1,  .949,  .8}60.1 & 68.9 \bigstrut\\
    \hline
    OCPL* & 76.0  & 88.6  & 80.6  & 66.1  & 69.2  & 85.2  & 88.6  & 89.0  & 56.4  & 86.7  & 74.4  & 86.1  & 88.7  & 87.6  & 85.4  & 50.1  & 83.7  & 74.9  & 77.8  & \cellcolor[rgb]{ 1,  .949,  .8}48.5 & 77.1 \bigstrut\\
    \hline
    BSDP(ORE*) & 89.5  & 89.3  & 82.6  & 69.8  & 72.8  & 89.6  & 92.3  & 91.7  & 64.6  & 87.5  & 74.2  & 88.6  & 90.7  & 87.0  & 88.1  & 57.8  & 85.9  & 79.9  & 84.6  & \cellcolor[rgb]{ 1,  .949,  .8}53.6 & 80.1 \bigstrut\\
    \hline
    BSDP(OCPL*) & 89.2  & 90.2  & 81.8  & 71.0  & 73.0  & 89.7  & 91.6  & 91.4  & 64.6  & 86.4  & 75.8  & 89.6  & 90.5  & 87.3  & 87.9  & 56.9  & 85.8  & 79.3  & 85.5  & \cellcolor[rgb]{ 1,  .949,  .8}57.0 & \textbf{81.2} \bigstrut\\
    \hline
    \end{tabular}}
  \label{table:3}
\end{table*}

It is obvious that our BSDP performed best. As the number of categories of the base task increases, our performance advantage is even greater, which is also consistent with the analysis mentioned in \hyperref[4.2.1]{Section 4.2.1}. Since the CIL experiments are performed offline, our data-level perturbation module helps less, leading to a little decrease in the performance of incremental categories. However, the effect of our feature-level perturbation module is enhanced by training more epochs, which results in less forgetting of the base categories. Generally, since BSDP performs well in the online manner, it obviously works better offline. Although our BSDP is designed for the OLOWOD problem, it still performs well on the iOD problem, which once again validates the outstanding ability of our dual-level perturbations to mitigate catastrophic forgetting. In the future, we will also consider how to modify our BSDP to improve the performance on incremental categories in an offline manner.

\subsubsection{Ablation study}  % 4.2.3
This section will provide more specific analyses of our BSDP approach through various ablation experiments. All experiments are carried out with the data splitting mentioned in \hyperref[4.1.1]{Section 4.1.1} and use 'OCPL*-online' as the baseline.

\paragraph{BSDP modules}~{}  % 4.2.3.1
\par
% 表4
\begin{table*}[width=1\textwidth, htbp]
  \centering
  \caption{The ablation study on the effect of different modules of our BSDP.}
  \setlength{\tabcolsep}{2.5mm}{
  \scalebox{0.81}{
    \begin{tabular}{c|c|c|c|c|c|c|c|c|c|c|c|c}
    \hline
    \multirow{2}[4]{*}{Methods} & \multicolumn{3}{c|}{Task 1 (mAP)} & \multicolumn{3}{c|}{Task 2 (mAP)} & \multicolumn{3}{c|}{Task 3 (mAP)} & \multicolumn{3}{c}{Task 4 (mAP)} \bigstrut\\
\cline{2-13}          & previous & current & both  & previous & current & both  & previous & current & both  & previous & current & both \bigstrut\\
    \hline
    baseline & /     & 56.15 & /     & 51.22 & 23.40 & 37.31 & 37.89 & 12.51 & 29.43 & 28.98 & 12.27 & 24.81 \bigstrut\\
    \hline
    PB   & /     & 56.15 & /     & 53.10 & 22.73 & 37.91 & 38.64 & 13.17 & 30.15 & 30.07 & 12.58 & 25.70 \bigstrut\\
    \hline
    PB+FLP & /     & 56.15 & /     & \textbf{54.05} & 22.42 & 38.23 & 38.99 & 13.33 & 30.42 & 30.58 & 12.66 & 26.10 \bigstrut\\
    \hline
    PB+DLP & /     & 56.15 & /     & 53.54 & 22.67 & 38.10 & 38.76 & \textbf{13.43} & 30.32 & 30.47 & 12.61 & 26.01 \bigstrut\\
    \hline
    PB+FLP+DLP & /     & 56.15 & /     & 53.28 & \textbf{23.73} & \textbf{38.50} & \textbf{39.31} & 13.37 & \textbf{30.66} & \textbf{30.73} & \textbf{12.88} & \textbf{26.26} \bigstrut\\
    \hline
    \end{tabular}}}
  \label{table:4}
\end{table*}

In order to better prove the effectiveness of each module of our BSDP, we conduct ablation experiments, and the results are shown in \hyperref[table:4]{Table 4}. The baseline in this table means 'OCPL*-online' and it used a random selection strategy. 'PB' represents the model that is composed of the baseline and our prototype-based sample selection strategy rather than the original random selection. 'FLP' stands for our feature-level perturbations module which perturb new category features with prototypes of old categories to preserve old knowledge. 'DLP' represents our data-level perturbations module using the statistics of old category distributions which aims to make the model learn the discriminative features for new categories and not forget old categories.

From the comparisons of the results from the first and second rows, it can be observed that the model with 'PB' has significantly improved performance on the previous categories(53.10\% vs. 51.22\% for task $T_2$, 38.64\% vs. 37.89\% for task $T_3$, and 30.07\% vs. 28.98\% for task $T_4$). This also results in better performance of the model detection on all known categories('both'). This confirms that the exemplars selected by our prototype-based sample selection strategy are more representative than those selected randomly. The second and third rows show that the mAPs of the previous categories have been promoted more than the mAPs of the current categories(0.95\% vs. -0.31\% for task $T_2$, 0.35\% vs. 0.16\% for task $T_3$, and 0.51\% vs. 0.08\% for task $T_4$). This proves that using prototypes of old categories to perturb new features does preserve old knowledge very well. And the performance of the new categories is also improved especially in tasks $T_3$ and $T_4$, which proves that the feature-level perturbations make good use of the correlation and discrimination between categories as we mentioned in \hyperref[3.4]{Section 3.4}. The results of the second and fourth rows prove that the generated data-level perturbations can indeed make use of the statistics of the old category distributions. Therefore, it can improve the mAPs of the new categories without catastrophic forgetting. By comparing the results of the first and fifth rows, we can easily see that the mAPs of all seen categories have a significant increase(1.19\% for task $T_2$, 1.23\% for task $T_3$, and 1.45\% for task $T_4$). It is obvious that our method improves the mAPs in both old and new categories significantly, which proves the effectiveness of our BSDP method.

\paragraph{Types of data-level perturbations}~{}  % 4.2.3.2
\par
% 表5
\begin{table*}[width=1\textwidth, htbp]
  \centering
  \caption{We carefully ablate some simple perturbation methods. All experiments are carried out with our prototype-based sample selection strategy by default.}
  \setlength{\tabcolsep}{2.4mm}{
  \scalebox{0.83}{
    \begin{tabular}{c|c|c|c|c|c|c|c|c|c|c|c|c}
    \hline
    \multirow{2}[4]{*}{Methods} & \multicolumn{3}{c|}{Task 1 (mAP)} & \multicolumn{3}{c|}{Task 2 (mAP)} & \multicolumn{3}{c|}{Task 3 (mAP)} & \multicolumn{3}{c}{Task 4 (mAP)} \bigstrut\\
\cline{2-13}          & previous & current & both  & previous & current & both  & previous & current & both  & previous & current & both \bigstrut\\
    \hline
    FLP   & /     & 56.15 & /     & \textbf{54.05} & 22.42 & 38.23 & 38.99 & 13.33 & 30.42 & 30.58 & 12.66 & 26.10 \bigstrut\\
    \hline
    FLP+aug & /     & 56.15 & /     & 53.49 & 22.98 & 38.24 & 38.96 & 13.29 & 30.41 & 30.64 & 12.66 & 26.15 \bigstrut\\
    \hline
    FLP+gaussian & /     & 56.15 & /     & 53.59 & 23.23 & 38.41 & 39.06 & 13.23 & 30.45 & 30.57 & \textbf{13.07} & 26.19 \bigstrut\\
    \hline
    FLP+DLP & /     & 56.15 & /     & 53.28 & \textbf{23.73} & \textbf{38.50} & \textbf{39.31} & \textbf{13.37} & \textbf{30.66} & \textbf{30.73} & 12.88 & \textbf{26.26} \bigstrut\\
    \hline
    \end{tabular}}}
  \label{table:5}
\end{table*}

To better highlight the effectiveness of our data-level perturbations in an online training manner, we compare it with general image transformation and adding general Gaussian noises, and the results are shown in \hyperref[table:5]{Table 5}. 'FLP' represents the model only uses feature-level perturbations. 'aug' represents the random general image transformation, such as brightness transformation, contrast transformation, saturation transformation, etc. 'gaussian' means adding Gaussian noises with mean 0 and variance 1 to the model. 'DLP' stands for our data-level perturbations. The frequency of all the perturbation methods is performed for 1\% training samples as mentioned in \hyperref[4.1.3]{Section 4.1.3}. By comparing the results of the second, third, and fourth rows, we can conclude that our data-level perturbations can improve the accuracy more effectively than other general methods. Since the model can only see samples once in an online training manner, the samples need to have more representative features. When training the model with 'aug', the original representative features may be replaced, resulting in performance degradation(30.41\% vs. 30.42\% for task $T_3$). Moreover, general Gaussian noises do not carry any information about the old categories, so the performance improvement on 'previous' is limited. In contrast, our data-level perturbation module makes better use of statistics from the old categories as old knowledge, which improves the mAPs of both the previous and the current categories.

\paragraph{Frequency of perturbations}~{}  % 4.2.3.3
\par
% 图片7
\begin{figure*}
	\centering
		\includegraphics[scale=.45]{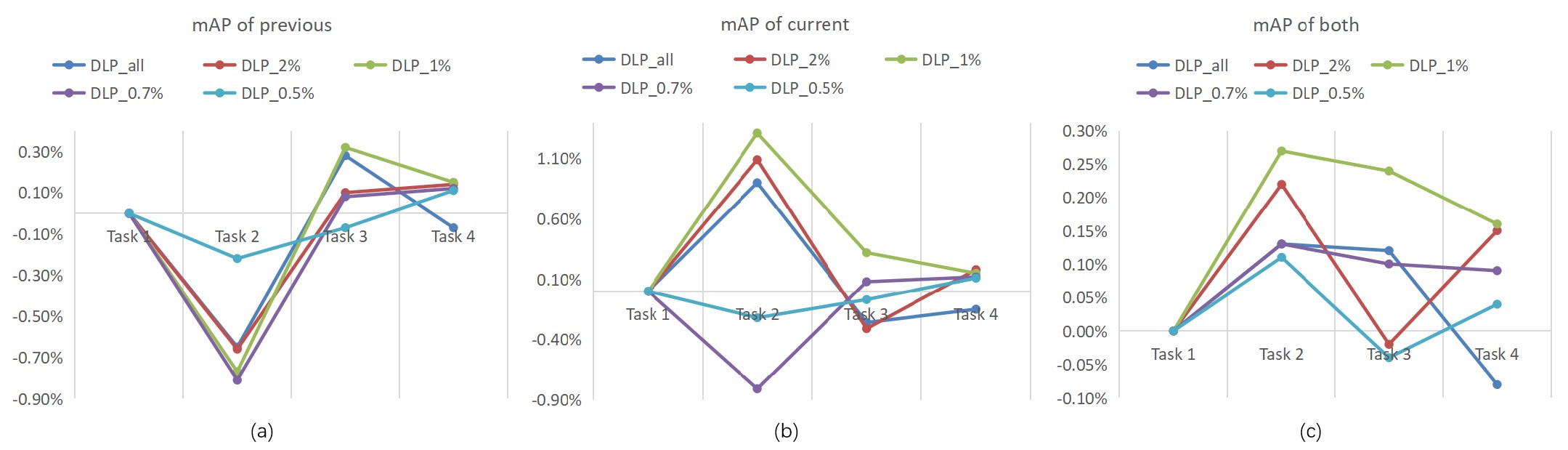}
	\caption{The rise of mAPs of previous, current, and all known(both) categories with different frequencies of data-level perturbations. 'DLP\_all' means that all samples are used to generate perturbations during training, while 'DLP\_2\%' indicates that data-level perturbation is performed for 2\% training samples, and so on.}
	\label{FIG:7}
\end{figure*}

We keep the other module unchanged and change the frequency of our data-level perturbation and the results are shown in \hyperref[FIG:7]{Figure 7}. It can be found that when performing 'DLP' for 1\% training samples, the mAPs of all seen('both') categories for all tasks are the best. The improvement in task $T_3$ is more pronounced than in other cases. Although 'DLP\_1\%' is dropped on the mAP of the previous categories in task $T_2$, it still performs best overall. It can be observed that the performance degradation in task $T_2$ is caused by the inclusion of more useful new categories information in the perturbed samples, as can be seen in \hyperref[FIG:7]{Figure 7(b)}. From the results of 'DLP\_0.5\%', it is obvious that when the frequency of data-level perturbation is reduced, the model's ability to learn new categories is worse(\hyperref[FIG:7]{Figure 7(b)}) but the performance degradation on the old categories are reduced. This proves that our data-level perturbation module is meaningful for learning new categories, and it can be found in tasks $T_3$ and $T_4$ that our method is also helpful for maintaining old knowledge(\hyperref[FIG:7]{Figure 7(a)}). Thus, a reasonable frequency can amplify the advantages of our approach.

% 表6
\begin{table}[htbp]
  \centering
  \caption{Ablation of the frequencies of perturbations. We only show the mAPs of 'both' for all tasks.}
  \setlength{\tabcolsep}{4.6mm}{
  \scalebox{0.78}{
  \normalsize
    \begin{tabular}{c|c|c|c|c}
    \hline
    Methods & Task 1 & Task 2 & Task 3 & Task 4 \bigstrut\\
    \hline
    FLP\_all + DLP\_1\% & 56.15 & \textbf{38.90} & \textbf{30.73} & \textbf{26.41} \bigstrut\\
    \hline
    FLP\_all + DLP\_all & 56.15 & 38.58 & 30.61 & 26.30 \bigstrut\\
    \hline
    FLP\_1\% + DLP\_1\% & 56.15 & 38.50 & 30.66 & 26.26 \bigstrut\\
    \hline
    FLP\_1\% + DLP\_all & 56.15 & 38.36 & 30.54 & 26.02 \bigstrut\\
    \hline
    \end{tabular}}}
  \label{table:6}
\end{table}

Interestingly, if we apply data-level perturbation to all training samples, the performance will drop dramatically. We do ablate experiments in the case of 'FLP\_all' and 'FLP\_1\%' respectively. As is shown in \hyperref[table:6]{Table 6}, 'DLP\_all' ranks last in all of the metrics compared with 'DLP\_1\%'. It also proves that in the case of limited samples, too many perturbed samples will disturb the original data distributions, which is not conducive to online training. So we choose 'DLP\_1\%' in our experiments. It is worth noting that the performance of the model is improved when the frequency of feature-level perturbation is increased, i.e. performed on each feature('FLP\_all'). Since 'FLP' helps to preserve old knowledge and improve the discrimination of the model without any negative effects, the performance of 'FLP\_all' is rightly better than that of 'FLP\_1\%'.

\paragraph{Quantity of stored features}~{}  % 4.2.3.4
\par
% 图片8
\begin{figure}
	\centering
		\includegraphics[scale=.62]{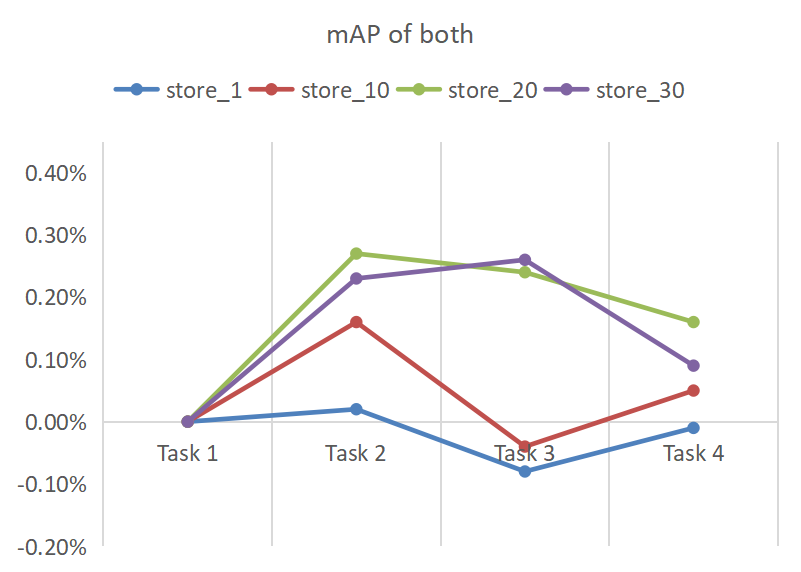}
	\caption{The influence of different numbers of stored features on distribution fitting of our data-level perturbation module. For convenience, we only show the results of mAPs of 'both'. 'store\_1' means we store only one group of features per category, i.e. 80 features for each category. store\_10' means to store 10 groups of features, and so on.}
	\label{FIG:8}
\end{figure}

Here we explore the effect of the quantity of features stored for each category on the distribution fitting of our data-level perturbation module. As shown in \hyperref[FIG:8]{Figure 8}, we show mAPs of all seen categories with different numbers of features. It is obvious that the fewer features we store, the less the generated perturbed samples contribute and even degrades performance at task $T_3$, because fewer features will affect the quality of distribution fitting and invalidate the generated perturbed samples. So when provided 20 groups of features, our data-level perturbation module works. Increasing the number of features more than 20 groups does little to improve performance and however, increases memory consumption. So we choose 'store\_20' in our experiments.

\paragraph{Distributions for fitting}~{}  % 4.2.3.5
\par
% 图片9
\begin{figure*}
	\centering
		\includegraphics[scale=.57]{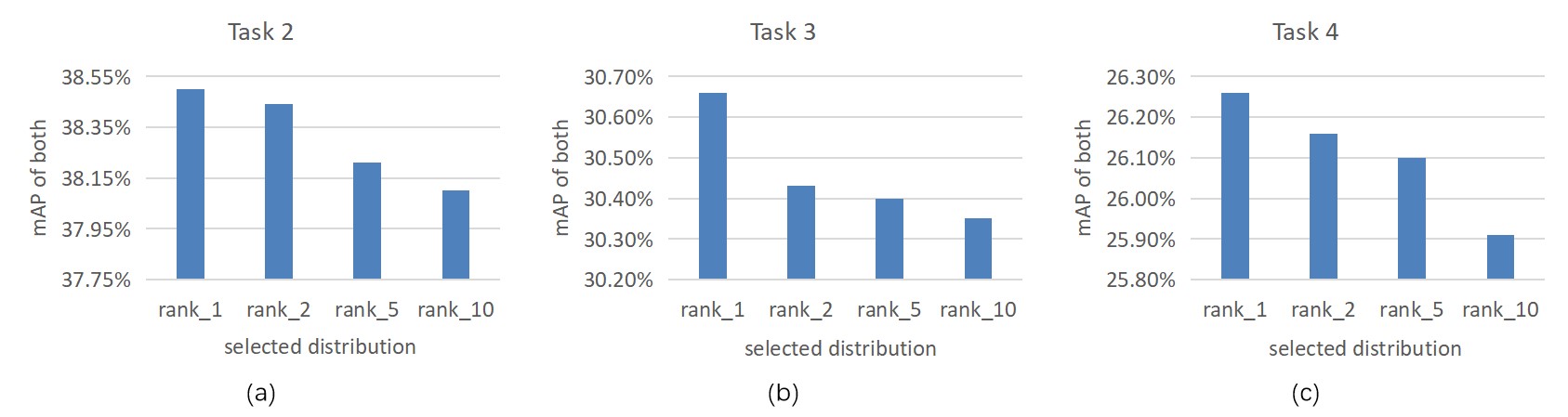}
	\caption{Ablation of the selected distribution in the data-level perturbation module. We use the stored features to fit 10 common distributions and then sort them according to the corresponding sum of the squared errors as described in \hyperref[3.5]{Section 3.5}. 'rank\_1' means that we choose the best-fitted distribution to generate perturbed samples. 'rank\_2' represents the sub-optimal distribution, and 'rank\_5' and 'rank\_10' represent the fifth-ranked and tenth-ranked(i.e. the worst-fitted) distributions, respectively.}
	\label{FIG:9}
\end{figure*}
% 图片10
\begin{figure*}
	\centering
		\includegraphics[scale=.29]{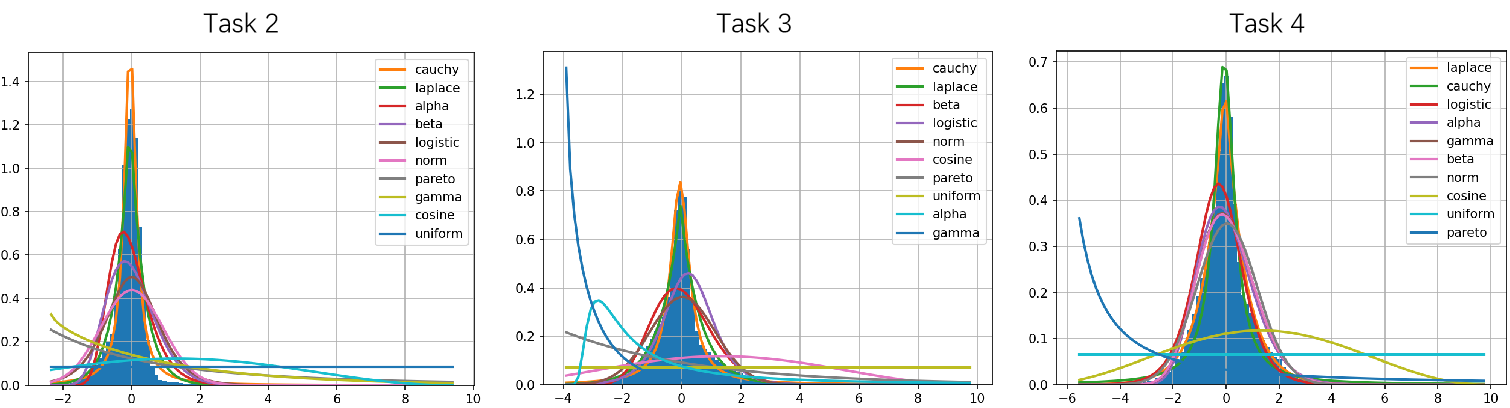}
	\caption{The results of fitting 10 common distributions with the stored features before training tasks $T_2$, $T_3$ and $T_4$, respectively.}
	\label{FIG:10}
\end{figure*}

We use the stored features to fit 'alpha', 'beta', 'gamma', 'laplace', 'uniform', 'norm', 'cauchy', 'logistic', 'pareto', and 'cosine' distributions. Before training each task(except task $T_1$), we fit the distributions with the features saved from the previous tasks.  The results of fitting distributions before tasks $T_2$, $T_3$ and $T_4$ are shown in \hyperref[FIG:10]{Figure 10}. It's worth mentioning that the best-fitted distribution varies from different tasks, which means the distribution we select varies from different tasks. In order to prove the validity of our proposed data-level perturbation module, we ablate the selected distribution to which the generated perturbations belong. The distributions of different ranks are ablated as shown in \hyperref[FIG:9]{Figure 9}. We present mAPs of all known categories of tasks $T_2$, $T_3$, and $T_4$. It is obvious that 'rank\_1' performs best, while the performance of 'rank\_2', 'rank\_5', and 'rank\_10' decline in order. It demonstrates the generated perturbed samples consolidate old knowledge and improve the robustness of the model.

\paragraph{Hyper-parameter $\gamma$}~{}  % 4.2.3.6
\par
% 表7
\begin{table*}[width=1\textwidth,htbp]
  \centering
  \caption{We ablate the hyper-parameter $\gamma$ of the feature-level perturbation module. We keep the other modules of BSDP(OCPL*) frozen and only change the hyper-parameter $\gamma$. The best results of 'both' have been bolded in this table.}
  \setlength{\tabcolsep}{1.9mm}{
  \scalebox{0.8}{
    \begin{tabular}{c|c|c|c|c|c|c|c|c|c|c|c|c|c}
    \hline
    \multirow{2}[4]{*}{Row ID} & \multirow{2}[4]{*}{Hyper-parameter $\gamma$} & \multicolumn{3}{c|}{Task 1 (mAP)} & \multicolumn{3}{c|}{Task 2 (mAP)} & \multicolumn{3}{c|}{Task 3 (mAP)} & \multicolumn{3}{c}{Task 4 (mAP)} \bigstrut\\
\cline{3-14}          &       & previous & current & both  & previous & current & both  & previous & current & both  & previous & current & both \bigstrut\\
    \hline
    1     & 0.4   & /     & 56.15 & /     & 54.10 & 21.74 & 37.93 & 38.99 & 13.05 & 30.34 & 30.57 & 12.58 & 26.03 \bigstrut\\
    \hline
    2     & 0.5   & /     & 56.15 & /     & 54.05 & 22.42 & \textbf{38.23} & 38.99 & 13.33 & \textbf{30.42} & 30.58 & 12.66 & \textbf{26.10} \bigstrut\\
    \hline
    3     & 0.6   & /     & 56.15 & /     & 53.29 & 23.05 & 38.17 & 38.52 & 13.32 & 30.12 & 30.34 & 12.38 & 25.85 \bigstrut\\
    \hline
    4     & 0.7   & /     & 56.15 & /     & 53.00 & 23.20 & 38.10 & 38.67 & 13.29 & 30.21 & 30.35 & 12.49 & 25.90 \bigstrut\\
    \hline
    \end{tabular}}}
  \label{table:7}
\end{table*}

In this part, we analyze our feature-level perturbation module without data-level perturbation. It is used to perturb features with prototypes of old categories to preserve old knowledge. The hyper-parameter $\gamma$ in \hyperref[Eq6]{Eq. 6} is used to adjust the weights of $f_{gen}$ and $f$. The larger the $\gamma$ is, the more the proportion of the features of the new categories have in $f_{pert}$. As shown in \hyperref[table:7]{Table 7}, our model can achieve the best results on 'both' of three incremental tasks for $\gamma =0.5$. According to task $T_2$, the mAPs of 'current' increases with the increase of $\gamma$, while the mAPs of 'previous' decreases at the same time, which proves that the generated feature-level perturbations can effectively preserve old knowledge and improve the discrimination of the model. For task $T_3$ and task $T_4$, the model parameters are affected by the previous online incremental learning tasks, so the trend on task $T_2$ has weakened in the last two tasks. Overall, the model can better balance the features of the new categories and the old categories with our feature-level perturbations, which promotes the model to learn better in an online manner.

\subsubsection{Visualization}  % 4.2.4
% 图片11
\begin{figure*}
	\centering
		\includegraphics[scale=.22]{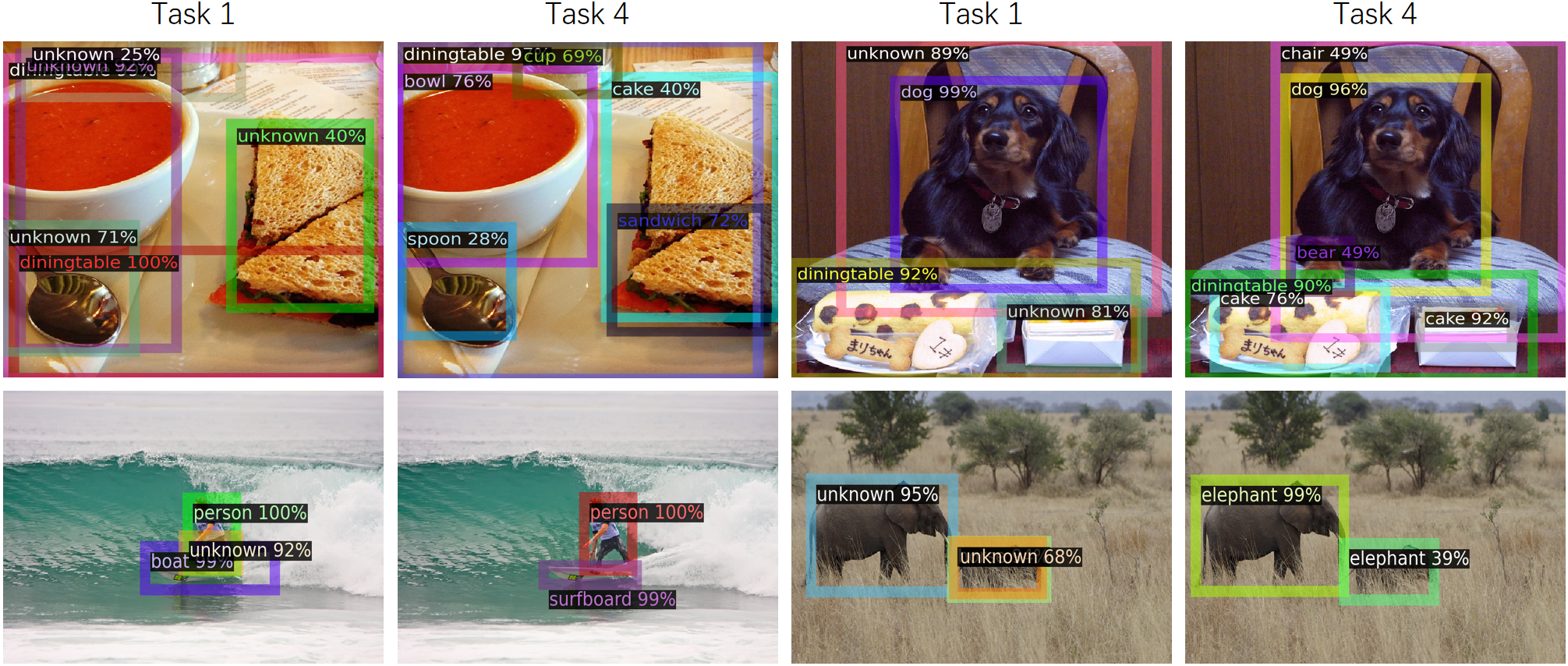}
	\caption{Visualization of our BSDP based on OCPL*. 'unknown' in the figure represents unknown categories.}
	\label{FIG:11}
\end{figure*}

To more intuitively demonstrate the effectiveness of our BSDP approach, we visualize some of the test results. As shown in \hyperref[FIG:11]{Figure 11}, we visualize the results of BSDP(OCPL*) after training on task $T_1$ and $T_4$. Columns 1 and 3 show the results of task $T_1$, while columns 2 and 4 show the results of task $T_4$. From \hyperref[FIG:11]{Figure 11}, it can be found that there are some unknown categories in task $T_1$ and they can be correctly detected in task $T_4$, such as 'spoon', 'sandwich', and 'cup' in Group 1 and 'cake', 'chair' in Group 3, and 'elephant' in group 4. Moreover, the accuracy of known categories is also improved. It verifies that our feature-level perturbation module can help the model make better use of the correlation and discrimination between categories to learn new knowledge and avoid forgetting. Therefore, the final model can not only identify the new categories but also maintain a high degree of confidence for the old categories, such as 'person' in Group 2 and 'dog' in Group 3.

% 图片12
\begin{figure*}
	\centering
		\includegraphics[scale=.22]{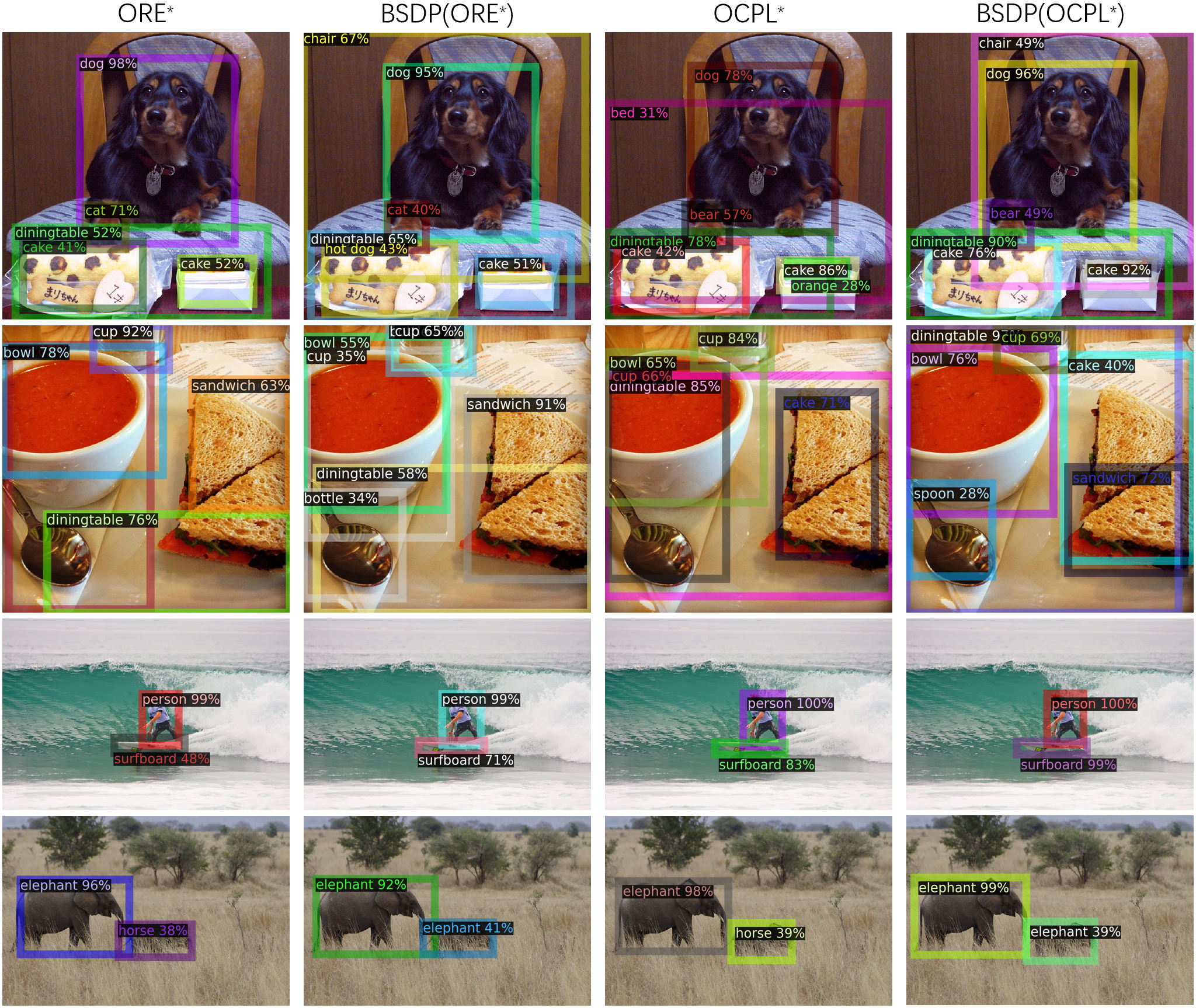}
	\caption{We visualize the results of task $T_4$ of both baselines and our BSDP approach. The columns 1 and 3 show the results of ORE* and OCPL*, and the columns 2 and 4 show the results of BSDP(ORE*) and BSDP(OCPL*), respectively.}
	\label{FIG:12}
\end{figure*}

In addition, we visualize both baselines and their BSDP versions to better observe the differences between the models and the advantages of our approach. The results of the visualization are shown in \hyperref[FIG:12]{Figure 12}. The columns 1 and 3 show the results of ORE* and OCPL*, and the columns 2 and 4 show the results of BSDP(ORE*) and BSDP(OCPL*), respectively.

Comparing the results of columns 1 with 2 and those of columns 3 with 4, our BSDP method has two advantages: (1) we improve the confidence of known categories and reduce wrong detections; (2) We detect more old categories that have been forgotten in ORE* and OCPL*. When comparing column 1 with 3 and column 2 with 4, it can be seen that OCPL* detects more targets with higher confidence than ORE* and the same for BSDP(OCPL*). Overall, BSDP(OCPL*) performs best. It can correctly detect 'cake', 'spoon', 'surfboard', and 'elephant' with high confidence.

\subsubsection{Limitations and future directions}  % 4.2.5
Although the proposed BSDP achieves good performance in tackling the OLOWOD problem, it also has some limitations at this stage. First, compared with offline learning, BSDP still has space for improvement. How to improve the performance of online training to match or even exceed offline training is one of our future directions. Second, BSDP focuses more on the data level and feature level, while not taking the model structure into account. In the future, we will pay more attention to modifying the model structure or making better use of multi-modal models. We will also take the large models into account and study the feasibility of combining our methods with those large models to tackle the OLOWOD problem better.

\section{Conclusions}  % 第五章 conclusions
In this paper, we first proposed the online open world object detection(OLOWOD) problem to simulate the learning manner of humans as they grow. The model needs to look at all the training data only once to incrementally learn multiple categories without forgetting previous categories. In addition, the model needs to identify 'unknown' and correctly detect the targets of these unknown categories in subsequent incremental tasks. We modified two OWOD methods as baselines for the OLOWOD problem. Then, we proposed a novel method called BSDP. Inspired by human brains, we suggest simulating the learning manner of humans by adding specific perturbations. Based on the memory-based incremental strategy, we selected the samples to replay according to the prototypes, and the selected samples are used to generate feature-level and data-level perturbations together with new category samples. The generated dual-level perturbations make good use of the correlation and discrimination between old and new categories and improve the robustness of the model effectively. Extensive experiments validated the good performance of our method.

Since the offline trained model can not cope with new tasks that can be frequently faced in the real world, we will put more effort into researching online training manner in the future. Training online can be much more helpful in real life and industry. Besides, we will focus on studying how to take advantage of multi-modal models to better improve the performance of the OLOWOD problem. We hope that this paper will contribute to the evolution of online incremental learning.

% \appendix
% \section{My Appendix}
% Appendix sections are coded under \verb+\appendix+.

% \verb+\printcredits+ command is used after appendix sections to list 
% author credit taxonomy contribution roles tagged using \verb+\credit+ 
% in frontmatter.

\printcredits

% \bibliographystyle{cas-model2-names}
\bibliographystyle{IEEEtran}

\end{sloppypar}
\end{document}